\documentclass[10pt]{article} 
\usepackage[accepted]{rlc}

\usepackage{amssymb}            
\usepackage{mathtools}          
\usepackage{mathrsfs}           
\usepackage{graphicx}           
\usepackage{subcaption}         
\usepackage[space]{grffile}     
\usepackage{url}                

\usepackage{amsmath}
\usepackage{amssymb}
\usepackage{mathtools}
\usepackage{amsthm}
\usepackage{bbm}
\usepackage{wrapfig}
\usepackage{colortbl}
\usepackage{multirow}

\usepackage[ruled,vlined]{algorithm2e}
\DontPrintSemicolon

\SetKwComment{Comment}{\color{green!50!black}// }{}

\newcommand{\FuncCall}[2]{\texttt{\bfseries #1(#2)}}

\SetKwProg{Function}{function}{}{}

\theoremstyle{plain}

\theoremstyle{definition}

\theoremstyle{remark}

\usepackage{amsmath,amsfonts,bm}

\def\eqref#1{equation~\ref{#1}}

\def\1{\bm{1}}

\def\vtheta{{\bm{\theta}}}
\def\va{{\bm{a}}}

\def\vg{{\bm{g}}}

\def\vs{{\bm{s}}}

\def\vx{{\bm{x}}}

\DeclareMathAlphabet{\mathsfit}{\encodingdefault}{\sfdefault}{m}{sl}
\SetMathAlphabet{\mathsfit}{bold}{\encodingdefault}{\sfdefault}{bx}{n}

\def\sA{{\mathcal{A}}}

\def\sD{{\mathcal{D}}}

\def\sG{{\mathcal{G}}}

\def\sP{{\mathcal{P}}}

\def\sR{{\mathcal{R}}}
\def\sS{{\mathcal{S}}}
\def\sT{{\mathcal{T}}}

\usepackage{soul}
\usepackage{xcolor}

\usepackage{nicefrac}

\title{Guided Data Augmentation for Offline Reinforcement Learning and Imitation Learning}

\author{Nicholas E. Corrado$^1$, Yuxiao Qu$^{2}$, John U. Balis$^1$, Adam Labiosa$^1$, Josiah P. Hanna$^1$ \\
University of Wisconsin--Madison$^1$, Carnegie Mellon University$^2$ \\
\texttt{\{ncorrado, balis, labiosa\}@wisc.edu, yuxaioq@andrew.cmu.edu, jphanna@cs.wisc.edu}
}

\begin{document}

\maketitle
\begin{abstract}

In offline reinforcement learning (RL), RL agents learn to solve a task using only a fixed dataset of previously collected data.
While offline RL has proven to be a viable method for learning real-world robot control policies, it typically requires large amounts of expert-quality data to learn effective policies that generalize to out-of-distribution states.
Unfortunately, such data is often difficult and expensive to acquire in real-world tasks.
Several recent works have leveraged data augmentation (DA) to inexpensively generate additional data, but most DA works apply augmentations in a random fashion and ultimately produce highly suboptimal augmented data.
In this work, we propose \textbf{Gu}ided \textbf{D}ata \textbf{A}ugmentation (GuDA), 
a human-guided DA framework that generates expert-quality augmented data.
The key insight behind GuDA is that while it may be difficult to demonstrate the sequence of actions required to produce expert data, a user can often easily characterize when an augmented trajectory segment represents progress toward task completion. 
Thus, a user can restrict the space of possible augmentations to automatically reject suboptimal augmented data.
To extract a policy from GuDA, we use off-the-shelf offline reinforcement learning and behavior cloning algorithms.
We evaluate GuDA on a physical robot soccer task as well as simulated D4RL navigation tasks, a simulated autonomous driving task, and a simulated soccer task.
Empirically, GuDA enables learning given a small initial dataset of potentially suboptimal experience and outperforms a random DA strategy as well as a model-based DA strategy.
%
We include videos and code at \url{https://nicholascorrado.github.io/projects/GuDA/}.
\end{abstract}

\section{Introduction}


%
%
Offline reinforcement learning (RL) is a learning paradigm in which RL agents learn to solve a task using only a static dataset of previously collected data.
While offline RL algorithms can produce effective real-world robot control policies without the expense or danger of active task interaction~\citep{levine2020offline}, their performance and generalization capabilities depend greatly on the size and quality of the provided dataset.
Ideally, we would provide large amounts of high-coverage, near expert-quality trajectories, but acquiring such data in real-world tasks is often challenging: the expense of data collection often limits us to just a few trajectories, and their quality depends on the performance of the data collection policy.
%
%
Although prior works have shown that offline RL algorithms can perform well even with highly suboptimal data~\citep{Kumar2019StabilizingOQ, fujimoto2019off, kumar2020conservative, fujimoto2021minimalist}, these same works show that these algorithm learn far more effective policies with expert-quality data.
As such, we focus on developing methods that produce expert-quality data without requiring a human to demonstrate expert behavior.

\begin{figure}
    \centering
    \includegraphics[width=\linewidth]{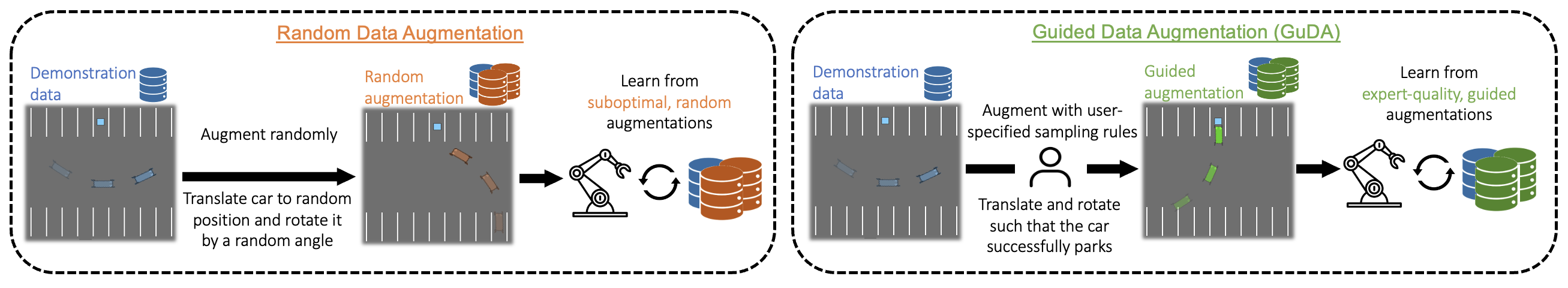}
    \caption{An overview of GuDA applied to a parking task given DAFs that translate and rotate a trajectory segment $\tau$.
    A user first defines a \textit{sampling procedure} describing how to translate and rotate $\tau$ to produce expert-quality data: translate $\tau$ so that the agent's final position is at the parking spot, and then rotate $\tau$ such that the agent is aligned with the parking spot.
    We augment our dataset using this sampling procedure and then learn a policy with offline RL or imitation learning.
    %
    %
    %
    %
    }
    \label{fig:guda_overview}
    \vspace{-0.5em}
\end{figure}

To improve performance and generalization of RL agents, a number of works have leveraged \textit{data augmentation}~\citep{laskin2020reinforcement} (DA), a technique in which agents generate additional synthetic experience without the expense of task interaction by applying transformations to previously collected experience.
%
%
%
These transformations -- or \textit{data augmentation functions} (DAFs) -- often leverage task-specific invariances and symmetries inherent to many real-world tasks (\textit{e.g.} translational invariance~\citep{pitis2020counterfactual, pitis2022mocoda}, gait symmetry~\citep{abdolhosseini2019learning, run_skeleton}).
Viewing DA as a means to improve dataset coverage, most prior works generate highly diverse augmented data by sampling data uniformly at random from a DAF~\citep{pmlr-v164-sinha22a, pitis2020counterfactual, joo2022swapping, cho2022S2PSI, Luetal20} or from a learned dynamics model~\citep{hepburn2022model, wang2022bootstrapped, 10003747}.
However, these random DA strategies generally produce highly suboptimal experience.
%
Thus, we aim to develop a DA strategy that produces both high-coverage and high-quality augmented data.
%
%

We propose \textbf{Gu}ided \textbf{D}ata \textbf{A}ugmentation (GuDA), a human-guided DA framework that generates large amounts of expert-quality data from a limited set of potentially suboptimal data.
The key insight behind GuDA is that a human can often determine if an augmented trajectory segment resembles expert data by simply checking if its sequence of states brings the agent closer to solving the task.
Thus, a user can restrict the space of DAF transformations to only generate augmented data that represents progress toward task completion.
%
%
%
To make this concept more concrete, imagine training an autonomous vehicle to park in a parking lot given a single suboptimal trajectory (Fig.~\ref{fig:guda_overview}).
Since a parking lot has a relatively uniform surface, we can generate augmented experience by translating and rotating the agent.
%
Sampling augmented data uniformly at random will most often produce data in which the agent drives away from the parking spot or approaches it at an unfavorable angle. 
However, we can generate expert-quality augmented data by translating and rotating trajectory segments such that the agent successfully parks. 
%


%
GuDA enables practitioners to generate expert data from potentially suboptimal experience without the expense of task interaction.
Additionally, instead of requiring that an expert provide an optimal sequence of actions solving a task, GuDA simply requires the user to characterize when an augmented trajectory segment represents progress toward task completion.
We evaluate GuDA with off-the-shelf offline RL algorithms on simulated navigation, autonomous driving, and soccer tasks as well as a physical robot soccer task. 
Since GuDA is also compatible with imitation learning algorithms (which require expert data), we also evaluate GuDA with behavior cloning.
%
%
Empirically, GuDA produces effective policies given a small amount of data -- even highly suboptimal data 
-- while a model-based DA strategy often fails due to poor model generalization. 
Moreover, polices trained under GuDA achieve larger returns than policies trained under a DA strategy that samples augmented data uniformly at random, emphasizing the importance of generating high-quality augmented data.
In summary, our core contributions are 
\begin{enumerate}
    \item We demonstrate how a human can guide data augmentation to inexpensively produce expert-quality data from potentially suboptimal experience. 
    \item We show that GuDA yields effective policies even when provided a small initial dataset.
    %
    \item We show that GuDA outperforms the most widely used DA strategy of sampling augmented data randomly, highlighting the benefits of generating expert-quality augmented data.
    %
    %
\end{enumerate}

\section{Related Work}

\subsection{Data Augmentation}
\label{sec:da}
Data augmentation (DA) refers to techniques that generate synthetic data by transforming previously collected experience and has been applied a variety of tasks, including algorithm discovery~\citep{fawzi2022discovering}, locomotion~\citep{run_skeleton, abdolhosseini2019learning}, and physical robot manipulation~\citep{george2023minimizing, mitrano2022data}.
%
%

%
DA is often used to generate perturbed data with the same semantic meaning as the original data.
Many vision-based RL works have trained agents to be robust to visual augmentations~\citep{laskin2020reinforcement, Guan2021WideningTP, wang2020improving, yarats2021mastering, raileanu2021automatic, hansen2021generalization, hansen2021stabilizing}, and similar approaches have been applied to non-visual tasks~\citep{sinha22a, weissenbacher2022koopman, qiao2021efficient}.
These approaches are orthogonal to GuDA; they use DA to improve policy robustness, while GuDA uses DA to improve dataset coverage and quality.
Perturbation-based DA methods more closely relate to domain randomization~\citep{sadeghi2016cad2rl, tobin2017domain, peng2018sim} which also aims for policy robustness.

Other works exploit invariances and symmetries in a task's dynamics to generate data that is semantically different from the original data.
Hindsight experience replay (HER)~\citep{andrychowicz2017hindsight, fang2018dher} counter-factually relabels a trajectory's goal.
Counterfactual Data Augmentation (CoDA)~\citep{pitis2020counterfactual} and Model-based CoDA (MoCoDA)~\citep{pitis2022mocoda} 
exploit local causal independence in a task's dynamics to generate additional data.
Several works use a learned model to generate augmented data~\citep{Luetal20, wang2022bootstrapped, hepburn2022model, sutton1990integrated, gu2016continuous, venkatraman2016improved, racaniere2017imagination}.
%
%
Most of these works focus on developing new DAFs and simply generate augmented experience in a random fashion.
In contrast, GuDA focuses on the importance of sampling expert-quality augmentations.

Two prior works closely relate to GuDA in that they aim to sample task-relevant augmented data: EXPAND~\citep{Guan2020WideningTP}, which applies visual augmentations to image regions identified by human feedback, and MoCoDA~\citep{pitis2022mocoda}, which generates augmented data by sampling $(\vs, \va)$ pairs from a user-defined \textit{parent distribution} $P(\vs, \va)$ and then computes $\vs'$ from a learned dynamics model.
%
%
%
GuDA differs from EXPAND in that GuDA focuses on non-visual tasks with DAFs more relevant to robotics. 
While MoCoDA can in principle generate expert data using an appropriately defined parent distribution, the user must specify the distribution over expert actions.
In contrast, GuDA requires no knowledge of the expert actions and simply requires the user to characterize data that represent task progress. 
Moreover, GuDA is a model-free DA framework and can be used when data is too scarce to model the task's dynamics, as is common in physical tasks.

\subsection{Offline Reinforcement Learning}

%
%
%
Offline RL~\citep{levine2020offline} methods learn a reward-maximizing policy from reward labels provided with a fixed dataset of task interactions.
These methods are designed such that, in principle, they can learn even with suboptimal data, though they are generally far more successful with expert data~\citep{Kumar2019StabilizingOQ, fujimoto2019off, kumar2020conservative, fujimoto2021minimalist}.

One core challenge with offline RL is extrapolation error: state-action pairs outside of the dataset's support can attain arbitrarily inaccurate state-action values during training, causing learning instabilities and poor generalization during deployment~\citep{gulcehre2020rl}.
This challenge is especially problematic for real-world robotics tasks in which offline data is scarce. 
Offline RL algorithms typically mitigate extrapolation error with policy parameterizations that only consider state-action pairs within the dataset~\citep{fujimoto2019off, ghasemipour2021emaq, zhou2021plas} or with behavioral cloning regularization~\citep{nair2020awac,fujimoto2021minimalist, xu2021offline}.
%
%
GuDA, like other DA strategies, can be viewed as a technique to mitigate extrapolation error by simply generating more data to improve dataset coverage without further task interaction. 
However, GuDA also improves dataset quality by generating expert-quality augmented data.

\section{Preliminaries}


\subsection{Offline Reinforcement Learning}


We consider finite-horizon Markov decision processes (MDPs)~\citep{puterman2014markov} defined by $(\sS, \sA, p, r, d_0, \gamma)$ where $\sS$ and $\sA$ denote the state and action space, respectively; $p(\vs' \mid \vs, \va)$ denotes the probability density of the next state $\vs'$ after taking action $\va$ in state $\vs$; and $r(\vs,\va)$ denotes the reward for taking action $\va$ in state $\vs$.\footnote{ 
If a reward function is unavailable, GuDA can be used with imitation learning methods such as behavior cloning which only assume access to expert data and do not require access to a reward function. }
We write $d_0$ as the initial state distribution, $\gamma \in [0, 1)$ as the discount factor,
and $H$ the episode length.
We consider stochastic policies $\pi_\vtheta : \sS \times \sA \to [0,1]$ parameterized by $\vtheta$.
The RL objective is to find a policy that maximizes the expected sum of discounted rewards $J(\vtheta) = \mathbb E\left[\sum_{t=0}^{H-1} \gamma^t r(\vs_t,\va_t)\right]$. 
In the offline RL paradigm, the agent cannot collect data through environment interaction and
must instead learn from a static dataset $\sD$ of transitions collected by a different policy.

\subsection{Data Augmentation Functions}

In this section, we introduce a general notion of a data augmentation function (DAF).
At a high level, a DAF generates augmented data by applying transformations to an input trajectory segment.
%
%
%
More formally, let $\sT$ denote the set of all possible trajectory segments and let $\Delta(\sT)$ denote the set of distributions over $\sT$. 
%
%
A DAF is a stochastic function $f: \sT \to \Delta(\sT)$ mapping a trajectory segment $((\vs_i, \va_i, r_i, \vs'_i))_{i=1}^k$ of length $k$ to an augmented trajectory segment $((\tilde\vs_i, \tilde\va_i, \tilde r_i, \tilde \vs'_i))_{i=1}^k$.
%
%
%
In this work, we focus on \textit{dynamics invariant} DAFs which produce realistic data that respect the task's dynamics and reward function, \textit{i.e.} $p(\tilde\vs' \mid \tilde\vs, \tilde\va) > 0$, and $\tilde r = r(\tilde\vs, \tilde\va)$ ~\citep{corrado2024understandingDA}.
%
%
As in most prior works, we assume a user can specify a DAF $f$ for a given domain~\citep{ pitis2020counterfactual}.


%
%
%
%
%

\begin{wrapfigure}{R}{0.42\linewidth}
    \vspace{-1em}
    \centering
    \includegraphics[width=\linewidth]{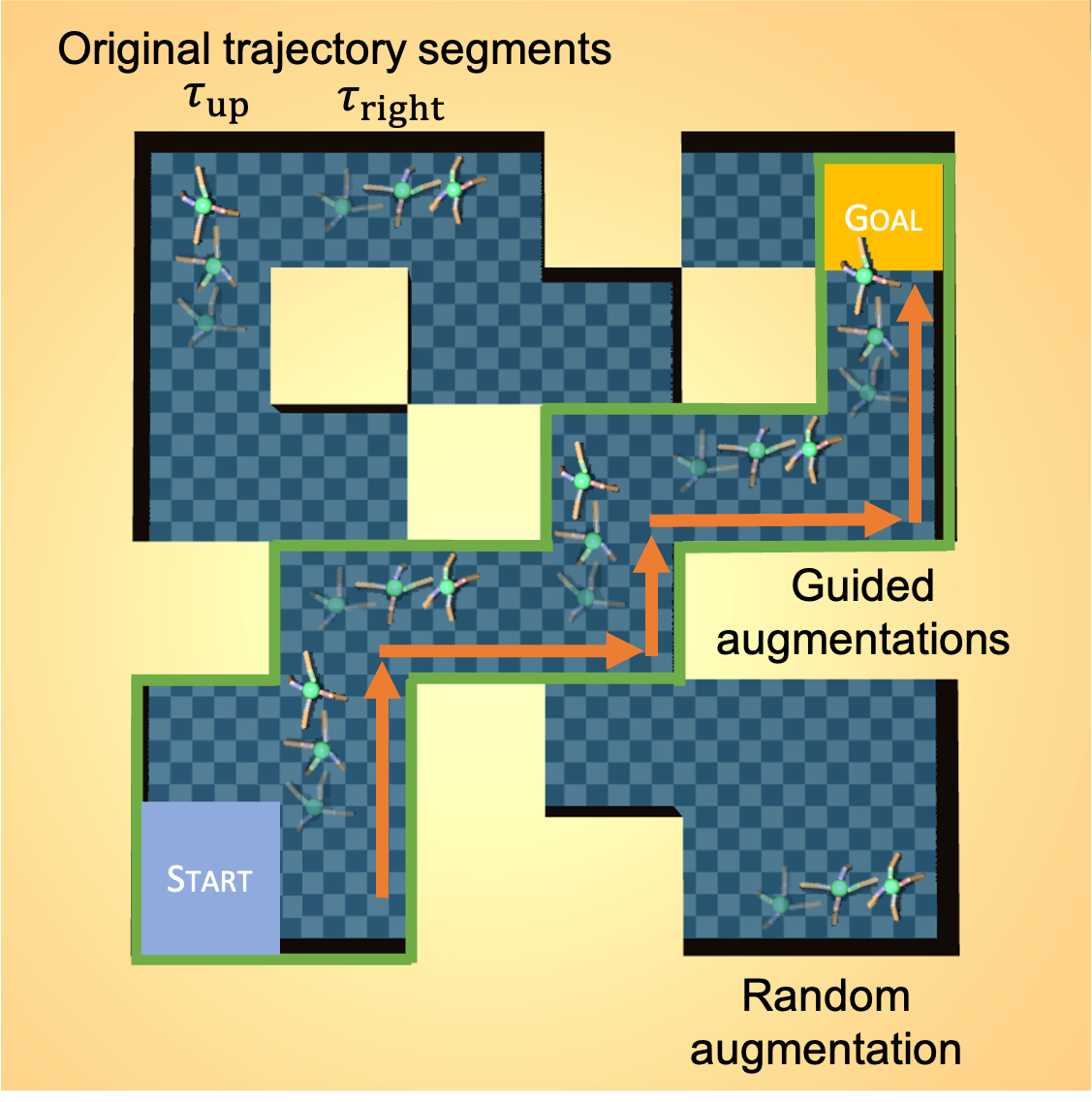}
    \caption{GuDA translates trajectory segments $\tau_\text{up}, \tau_\text{right}$ to demonstrate the agent walking to the goal. A random translation (bottom right) may be highly suboptimal. 
    }
    \label{fig:example_antmaze}
    \vspace{-1.5em}
\end{wrapfigure}

\section{Guided Data Augmentation}

In this section, we introduce Guided Data Augmentation (GuDA), a DA framework that automatically generates expert-quality augmented data.
We provide a high-level overview of GuDA in Section~\ref{sec:method_overview}, and then describe how we implement GuDA in Section~\ref{sec:implementation}.

\subsection{Method Overview}
\label{sec:method_overview}

%
We assume access to a dataset $\sD$ of task interactions and DAFs $f_1, \dots, f_m$. 
%
Prior to offline training, GuDA generates an augmented dataset $\widetilde \sD$ consisting of the original dataset plus $n$ augmented samples generated from the composition of DAFs $f = f_1 \circ \cdots \circ f_m$.
Afterwards, an agent learns from $\widetilde\sD$ using an off-the-shelf offline RL or imitation learning algorithm.
The core difference between GuDA and previous DA works lies in how GuDA samples augmented data from $f$.
Prior works typically sample augmented data uniformly at random, but most transformations under $f$ produce highly suboptimal experience.
However, a user can often easily characterize when an augmented trajectory segment represents progress toward task completion.
%
%
Thus, to generate augmented data that closely resembles expert data, GuDA has the user define a  \textit{sampling procedure} that describes how to sample augmentations from $f$ to produce data in which the agent makes task progress.

To illustrate how a user might identify a sampling procedure, consider a maze navigation task in which a  legged robot must reach a fixed goal state from a fixed initial position (Fig.~\ref{fig:example_antmaze}). 
We assume access to a DAF that translates the agent to a new position.
While it is difficult to demonstrate the precise sequence of leg movements required to optimally solve the maze, we can easily identify when a trajectory segment progresses the agent toward its goal.
A randomly sampled augmentation from our DAF will most likely have the agent visit maze regions that an expert would never visit and may even show the agent moving \textit{away} from the goal rather than toward it.
To ensure we generate expert augmented data, we can simply restrict our DAF to only sample new positions near the shortest path to the goal (green region) for which the agent's displacement is closely aligned with the shortest path (orange arrows).
This approach shifts the burden from the user having to demonstrate optimal actions to the user simply having to understand when augmented data represent progress toward task completion.
In the next section, we describe the DAFs we use and the sampling procedures we define to  generate augmented data that shows task progress.

%



\subsection{Implementation}

\label{sec:implementation}

GuDA's sampling procedures are domain-specific and depend on which DAFs are available as well as what task progress looks like in a given domain.
%
%
%
%
In this work, we consider four DAFs that transform an input trajectory segment $\tau$ using invariances and symmetries common to many physical tasks:
\begin{enumerate}
    \item \textbf{\texttt{Translate}$(\tau; \sP)$:} 
    Since the dynamics of agents and objects are often independent of their position, we can translate them to a new position $(x, y)$ sampled from a distribution $\sP$.
    %
    %
    \item \textbf{\texttt{Rotate}$(\tau; \Theta)$:} 
    Since the dynamics of agents and objects are often independent of their orientation, we can rotate the direction the agent and/or object faces by an angle $\theta$ sampled from a distribution $\Theta$ to produce motion in a different direction.
    %
    %
    \item \textbf{\texttt{Reflect}$(\tau; \sR)$:} 
    An agent that moves to the left often produces a mirror image of an agent moving to the right, so we can reflect the agent's left-right motion with probability $\sR(\tau)$.
    \item \textbf{\texttt{RelabelGoal}$(\tau; \sG)$:} In goal-conditioned tasks, dynamics are generally independent of the desired goal state~\citep{andrychowicz2017hindsight}.  
    Thus, we can replace the true goal with a new goal $\vg$ sampled from the task's goal distribution $\sG$.
    %
\end{enumerate}

\begin{wrapfigure}{R}{0.47\linewidth}
\vspace{-1.5em}
\begin{algorithm}[H]
    $\sG \gets$ distribution over task-relevant goals.\; 
    $\sP(x,y | \tau) \gets$ distribution over task-relevant positions for trajectory segment $\tau$.\;
    $\Theta(\theta | \tau) \gets$ distribution over task-relevant rotation angles for trajectory segment $\tau$.\; 
    $\sR(\tau) \gets$ probability of reflecting $\tau$.\;
  \Function{GuidedDAF($\tau_0$)}{
    $\tau \gets \texttt{copy}(\tau_0)$\;
    $\tau \gets$ \FuncCall{RelabelGoal}{$\tau; \sG$}\;
    $\tau \gets$ \FuncCall{Translate}{$\tau; \sP(x,y|\tau)$}\;
    $\tau \gets$ \FuncCall{Reflect}{$\tau; \sR(\tau)$}\;
    $\tau \gets$ \FuncCall{Rotate}{$\tau; \Theta(\theta|\tau)$}\;
    \For{$(\vs, \va, r, \vs') \in \tau$}{ 
        $r \gets r(\vs, \va)$  \Comment{Recompute rewards}}
    \Return{$\tau$}
  }
\caption{Guided Data Augmentation}
\label{alg:guided_daf}
\end{algorithm}
\vspace{-1em}
\end{wrapfigure}

We focus on navigation and manipulation tasks which have intuitive notions of task progress:
an agent makes progress if it moves closer to a goal position (navigation) or if it moves an object closer to a goal position (manipulation).
%
%
Given these notions of task progress, the user must specify how to apply these DAFs to generate expert-quality augmented data.
Formally, the user specifies distributions over translations $\sP(x, y|\tau)$, rotations $\Theta(\theta|\tau)$, and/or reflections $\sR(\tau)$ that produce data showing task progress.
%
%
To provide a concrete example of one such distribution, we return to the quadruped maze example in Fig.~\ref{fig:example_antmaze}. 
%
A human can easily identify task-relevant maze positions $(x, y)$ (green region) and a near-optimal displacement directions $\theta^*(x, y)$ for these positions (orange arrows).
Thus, to generate expert-quality augmented data using only the \texttt{Translate} DAF, we can sample new positions from $\sP(x, y|\tau) = \textsf{Unif}(\{(x, y)  : |\theta(\tau) - \theta^*(x,y)| \leq \frac{\pi}{4} \text{ and } (x, y) \text{ is within the green region}\})$, a uniform distribution over task-relevant maze positions for which the agent's original displacement direction $\theta(\tau)$ is closely aligned with $\theta^*(x, y)$.
If $\sP(x,y|\tau)$, $\Theta(\theta|\tau)$, and $\sR(\tau)$ are uniform distributions independent of $\tau$ over all valid position, rotations, and reflections, then GuDA reduces to the standard DA strategy that samples augmented data uniformly at random.

Algorithm~\ref{alg:guided_daf} provides pseudocode for our implementation of GuDA assuming access to all four DAFs.\footnote{GuDA can be implemented in many different ways and can be adapted depending on which DAFs are available. 
For instance, it is possible to guide DA by applying a subset of these four DAFs in a different order.
%
}
Table~\ref{tab:guided_augmentations} describes high-level sampling procedures for tasks in our empirical analysis: D4RL maze2d and antmaze navigation tasks~\citep{fu2020d4rl}, a parking task~\citep{highway-env}, a simulated robot soccer task, and a physical robot soccer task. 
Assuming access to all DAFs, the sampling procedures generally proceed as follows.
First, we randomly sample a new goal.
If $\tau$ is a full trajectory, we \texttt{Translate} $\tau$ such that the agent or object's final position is at the goal, and then \texttt{Reflect} and/or \texttt{Rotate} $\tau$ randomly about the goal.
If $\tau$ is a partial trajectory, we \texttt{Translate} $\tau$ to a new position that would likely be observed by an expert policy, and then 
\texttt{Reflect} and/or \texttt{Rotate} $\tau$ so that the agent or object moves as close as possible to the goal.
We provide task descriptions in Appendix~\ref{app:task_descriptions} and a more formal description of our sampling procedures in Appendix~\ref{app:sampling_rules}.

\section{Experiments}

We design an empirical study to evaluate two core hypotheses:
\begin{enumerate}
    \item[\textbf{H1:}] GuDA enables learning from a small dataset of potentially suboptimal data.
    \item[\textbf{H2:}] GuDA yields larger returns than a random DA strategy.

\end{enumerate}

\textbf{H1} implies that GuDA is well-suited to offline learning for real-world tasks where expert data is often scarce, and \textbf{H2} emphasizes the importance of sampling expert-quality augmented data. 
We note that support for \textbf{H2} implicitly provides support for \textbf{H1}.

%
%
%

%
%

%


\subsection{Simulated Experiments}


\begin{table*}[]
    \centering
    \begin{tabular}{r||c||l}
         & \textbf{Initial} & \multicolumn{1}{c}{\textbf{GuDA Sampling Procedures}}\\ 
         \multicolumn{1}{c||}{\textbf{Task Name}} & \textbf{Dataset Size} & \multicolumn{1}{c}{\textbf{($\tau$ = input trajectory segment)}}\\
         \hline
         \hline
         \multirow{3}{2.7cm}{maze2d-umaze maze2d-medium maze2d-large} 
         & \multirow{3}{2.2cm}{5 trajectories 5 trajectories 5 trajectories} &
         \multirow{3}{8.6cm}{\texttt{Translate} a partial trajectory $\tau$ to a random maze position, and then \texttt{Rotate} $\tau$ such that the agent moves along the shortest path to the goal. 
         See Fig.~\ref{fig:guda_maze2d}.
         }\\ 
         & & \\ 
         & & \\ 

         \hline         
         \multirow{3}{2.7cm}{antmaze-umaze antmaze-medium antmaze-large} 
         & \multirow{3}{2.2cm}{1 trajectory \\ 2 trajectories 5 trajectories} & \multirow{3}{8.6cm}{\texttt{Translate} a partial trajectory $\tau$ to a random maze position such that the agent moves along the shortest path to the goal. See Fig.~\ref{fig:example_antmaze}.}   \\ 
          & &  \\ 
          & & \\ 

         \hline
         \multirow{5}{2.7cm}{parking} & 
         \multirow{5}{2.2cm}{10 trajectories} &
         \multirow{5}{8.6cm}{Here, $\tau$ is a \textit{full trajectory}. First, use \texttt{RelabelGoal} to change $\tau$'s goal to randomly sampled goal (parking spot). Then, \texttt{Translate} $\tau$ such that the agent's final position is at the goal, and \texttt{Rotate} $\tau$ such that the car is within the parking spot. See Fig.~\ref{fig:guda_overview}.} \\
         & & \\ 
         & & \\ 
         & & \\ 
         & & \\ 
         
         \hline 
         \multirow{3}{2.7cm}{soccer-sim} & 
         \multirow{3}{2.2cm}{3 trajectories} &
         \multirow{3}{8.6cm}{Here, $\tau$ is a \textit{full trajectory}. \texttt{Reflect} $\tau$ with probability 0.5, \texttt{Translate} $\tau$ such that the ball's final position is at the goal, and then \texttt{Rotate} $\tau$ randomly.  
         See Fig.~\ref{fig:guda_soccer_sim}.
         } \\ 
         & & \\
         & & \\ 
         \hline
          \multirow{1}{2.7cm}{soccer-physical} & 
         \multirow{1}{2.2cm}{1 trajectory} &
         \multirow{1}{8.6cm}{See Section~\ref{sec:physical}.
} \\

    \end{tabular}
    \caption{GuDA sampling procedures for tasks in our empirical analysis. We provide task descriptions in Appendix~\ref{app:task_descriptions} and describe how we implement these sampling procedures in Appendix~\ref{app:sampling_rules}.
    }
    \label{tab:guided_augmentations}

\end{table*}

\begin{figure*}
    \centering
    \includegraphics[width=\linewidth]{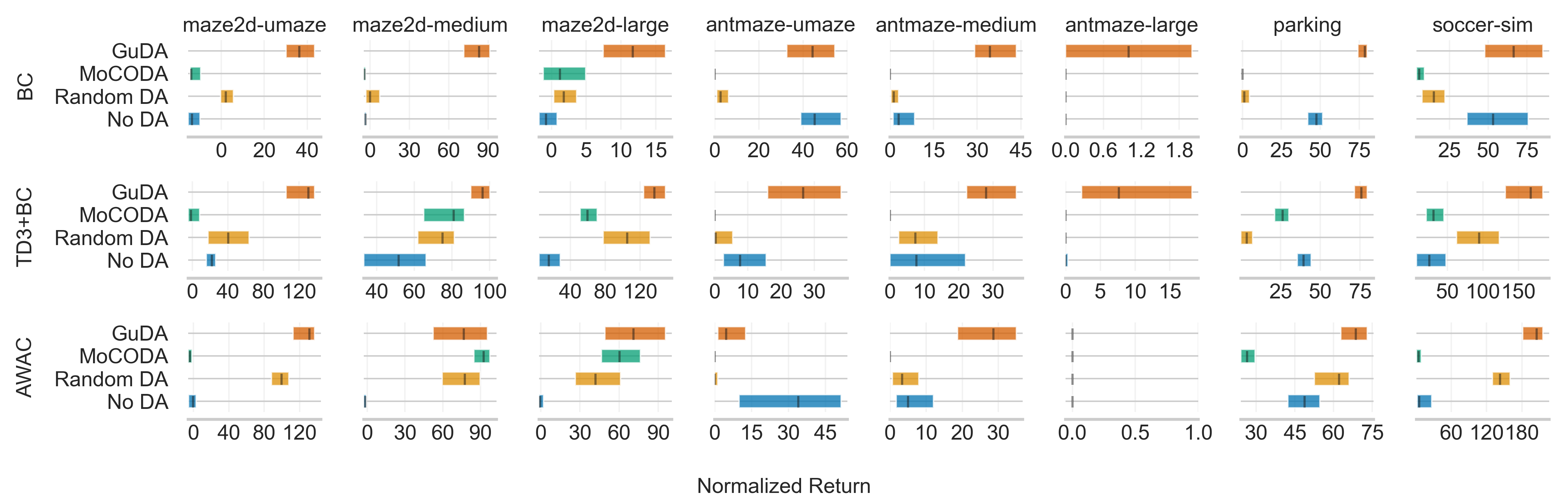}
    \caption{IQM normalized returns over 10 independent runs with 95\% stratified bootstrap confidence intervals for different DA strategies and algorithms.
    We compute normalized returns computed as $= 100 \cdot \frac{R - R_\text{random}}{R_\text{expert} - R_\text{random}}$ where $R_\text{expert}$ and $R_\text{random}$ denote the average return of the demonstrator and a policy that chooses actions uniformly at random, respectively, computed over 100 trajectories.
    }
    \label{fig:results}
    \vspace{-1em}
\end{figure*}

We first evaluate GuDA on simulated tasks described in Table~\ref{tab:guided_augmentations}.
In all tasks, we start with a small initial dataset containing at least one successful -- though not necessarily expert-level -- trajectory (Table~\ref{tab:guided_augmentations}).
These datasets contain failures and suboptimal behaviors as well: maze2d datasets contain data in which the agent moves away from the goal, soccer datasets contain trajectories where the agent kicks the ball out of bounds, and parking datasets contain trajectories where the car fails to park.
For maze2d and antmaze tasks, we hand-pick a small number of trajectory segments from the original `-v1' and `-diverse-v1' D4RL datasets, respectively.
For the remaining tasks, we use pre-trained policies to generate datasets. 
Dataset visualizations can be found in Fig.~\ref{fig:initial_datasets} of Appendix~\ref{app:task_descriptions}.

\begin{wrapfigure}{R}{0.4\linewidth}
    \vspace{-1em}
    \begin{subfigure}{0.49\linewidth}
        \centering
        \includegraphics[width=\linewidth]{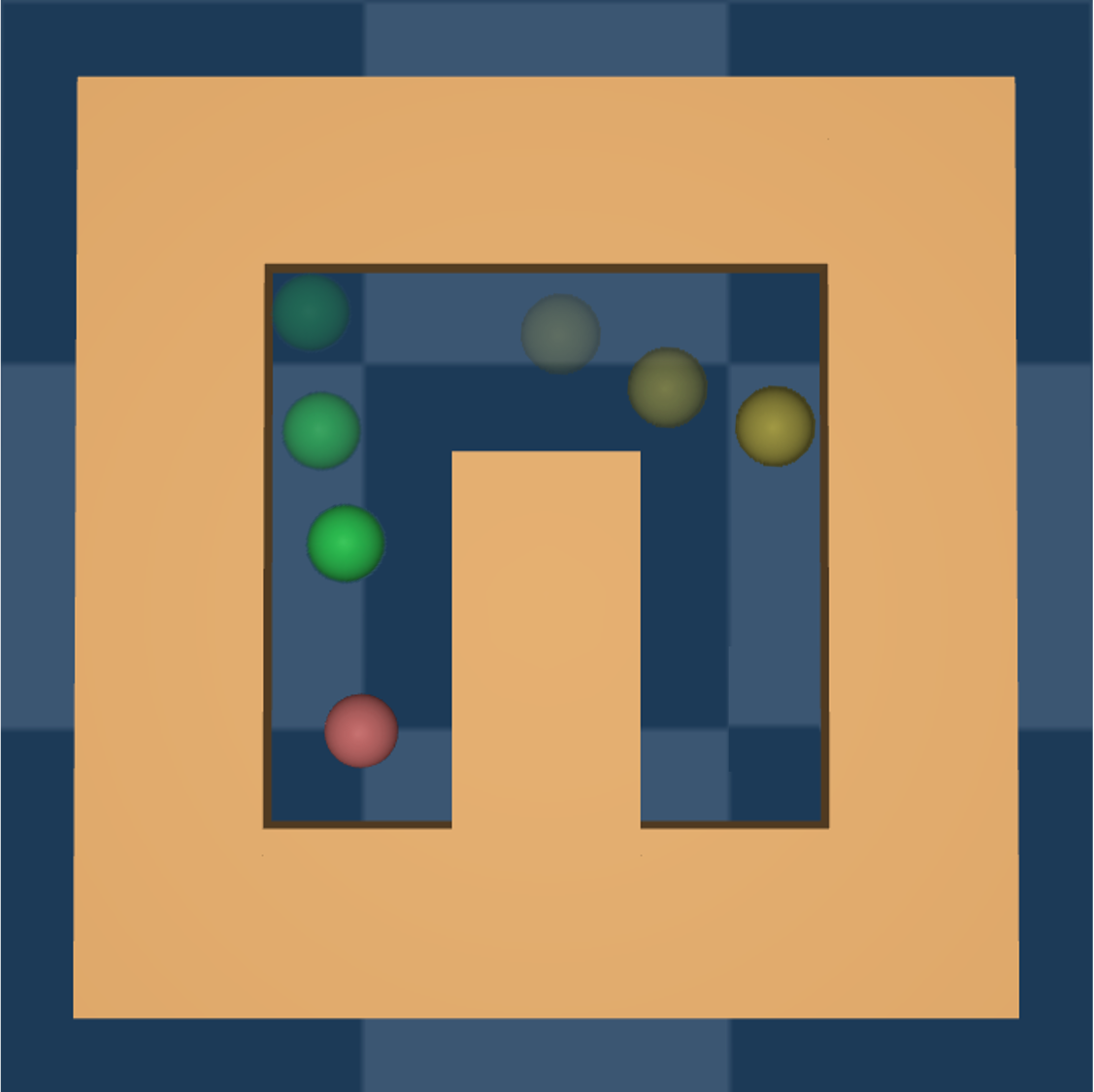}
        \caption{maze2d}
        \label{fig:guda_maze2d}
    \end{subfigure}
    \hfill
    \begin{subfigure}{0.49\linewidth}
        \centering
        \includegraphics[width=\linewidth]{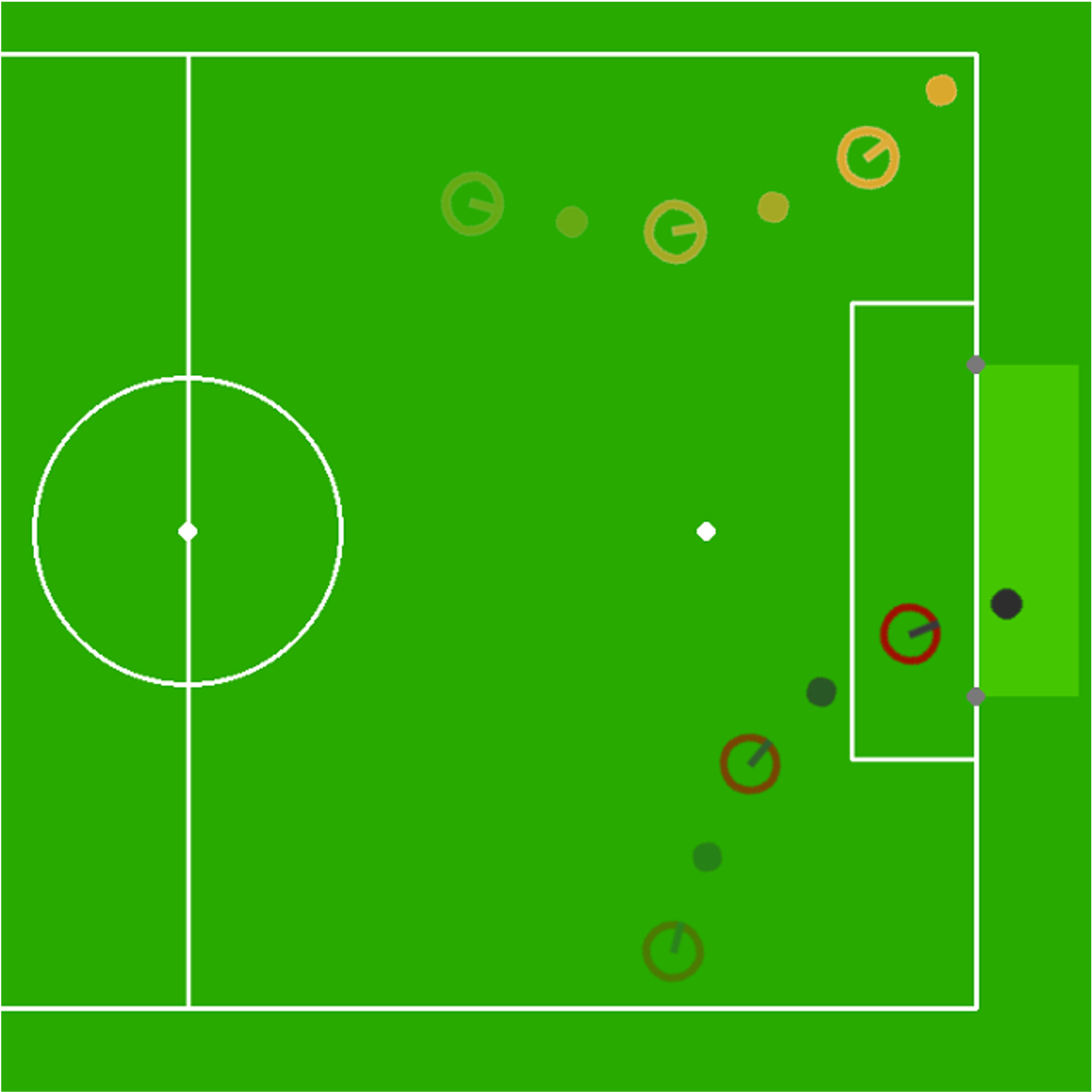}
        \caption{soccer-sim}
        \label{fig:guda_soccer_sim}
    \end{subfigure}
    \caption{Example augmentations under GuDA. The original trajectory segment is shown in yellow.}
    \label{fig:guided_dafs}
\end{wrapfigure}

We consider three baselines: the model-based DA strategy MoCoDA~\citep{pitis2022mocoda}, a DA strategy that randomly samples augmented data (Random DA), and no augmentation (No DA).
MoCoDA is a well-suited model-based baseline for our experiments; it exploits causal independence in the task's dynamics to efficiently learn a dynamics model that generalizes outside of the support of the dataset, which is particularly important when data is scarce. 
To improve the quality of MoCoDA data, we sample augmented states from a \textit{parent distribution} that closely matches the distribution of augmented states under GuDA.\footnote{Since we cannot identify expert state-action pairs, we only specify a parent distribution over task-relevant states.}
%
%
%
%
%
%
%
%
%
%
%
We provide further details on how we apply MoCoDA to each task in Appendix~\ref{app:mocoda}.
With each DA strategy, we generate 1 million augmented transitions and then perform offline learning with BC, TD3+BC~\citep{fujimoto2021minimalist}, and AWAC~\citep{nair2020awac} for 1 million policy updates. 
We tune hyperparameters for each algorithm and DA strategy separately using a hyperparameter sweep described in Appendix~\ref{app:hyperparams}.
We report the inter-quartile mean (IQM) return with 95\% bootstrap confidence intervals over 10 independent runs.\footnote{We choose to report the IQM because it is less biased and more statistically efficient than the median, and it is more robust to outliers than the mean~\citep{agarwal2021deep}.}

Fig.~\ref{fig:results} shows IQM normalized returns for each algorithm in each task.
GuDA almost always outperforms all baselines -- and often by a large margin (supporting \textbf{H1}).
For instance, in antmaze-medium, GuDA yields returns 3x larger than the next best strategy for all algorithms.
GuDA with TD3+BC is also the only strategy that can solve antmaze-large with significance.
Moreover, we emphasize that BC often achieves much larger returns with GuDA than with Random DA or MoCoDA, indicating that GuDA indeed generates expert data.
MoCoDA is unable to solve the more complex antmaze, parking, and soccer-sim tasks because it does not have enough data to learn an accurate, generalizable dynamics model, emphasizing the utility of GuDA in data-scarce settings.

While Random DA is often beneficial in maze2d and soccer-sim tasks, it often performs \textit{worse} than No DA in other tasks.
For instance, Random DA harms performance with all algorithms in antmaze-umaze, with BC and AWAC in antmaze-medium, and with BC and TD3+BC in parking.
Since BC mimics the provided data, it is understandable that Random DA may harm performance with BC.
However, since offline RL algorithms can learn from suboptimal data, these findings emphasize the importance of generating expert augmented data even for offline RL (supporting \textbf{H2}).

%


%
%
%
%


\subsection{Physical Experiments}
\label{sec:physical}

\begin{figure*}[t]
    \centering
    \begin{subfigure}{0.23\linewidth}
        \centering
        \includegraphics[width=\linewidth]{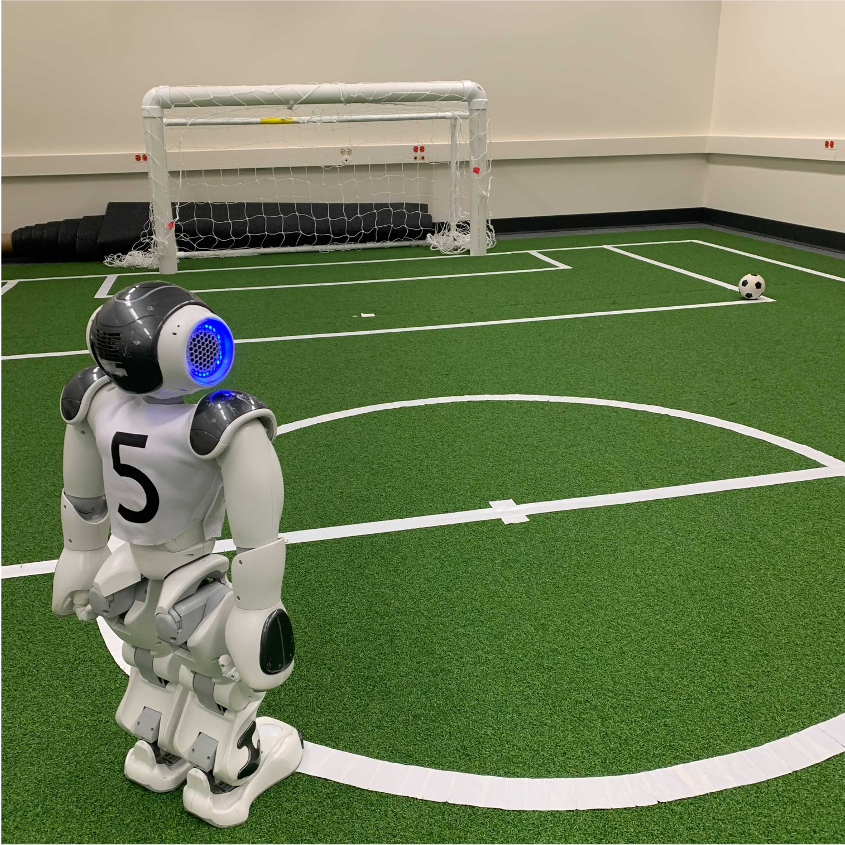}
        \caption{Easy initialization}
        \label{fig:init_1}
    \end{subfigure}
    \hfill
    \begin{subfigure}{0.23\linewidth}
        \centering
        \includegraphics[width=\linewidth]{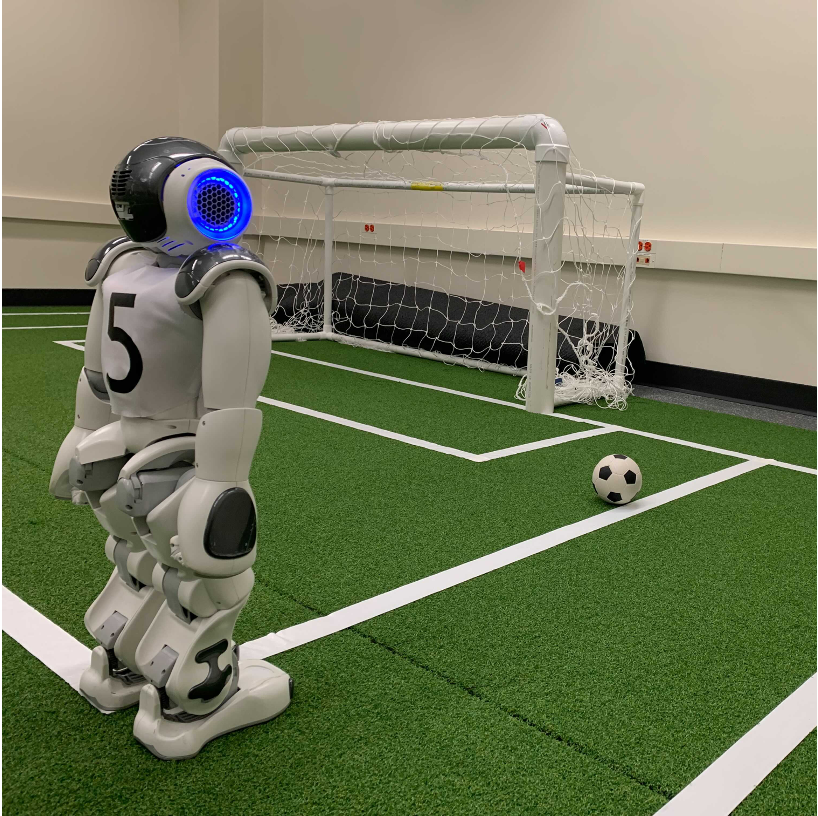}
        \caption{Hard initialization}
        \label{fig:init_2}
    \end{subfigure}
    \hfill
    \begin{subfigure}{0.23\linewidth}
        \centering
        \includegraphics[width=\linewidth]{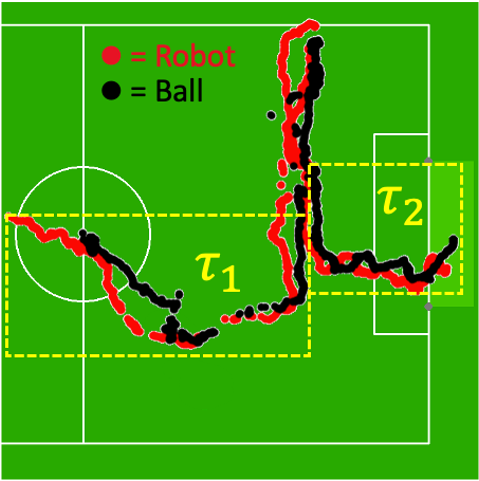}
        \caption{Initial dataset}
        \label{fig:demo}
    \end{subfigure}
    \hfill
    \begin{subfigure}{0.23\linewidth}
        \centering
        \includegraphics[width=\linewidth]{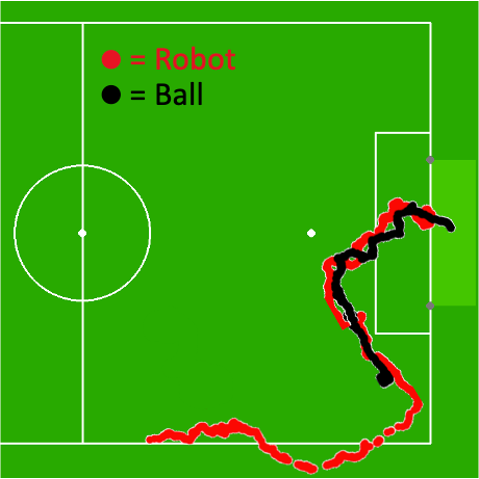}
        \caption{Example GuDA data}
        \label{fig:guda_demo}
    \end{subfigure}
    \caption{(\ref{fig:init_1}, \ref{fig:init_2}) Task initializations. (\ref{fig:demo}) Initial data with relevant segments $\tau_1$ and $\tau_2$. (\ref{fig:guda_demo}) An illustration of GuDA data generated by translating, rotating, and/or reflecting $\tau_1$ and $\tau_2$.}
    \label{fig:soccer_data}
\end{figure*}


\begin{wraptable}{R}{0.349\linewidth}
    \begin{tabular}{r||c|c}
         \textbf{Method} &  \textbf{Easy} & \textbf{Hard} \\
         \hline
         \multirow{1}{*}{\textbf{GuDA}} & 8/10 & \cellcolor{green!25}7/10 \\ 
         \multirow{1}{*}{\textbf{MoCoDA}} & 0/10 & 0/10  \\
         \multirow{1}{*}{\textbf{Random DA}} & 4/10 & 0/10\\
         \multirow{1}{*}{\textbf{No DA}} & 4/10 & 0/10\\
         \multirow{1}{*}{\textbf{Demonstrator}} & 9/10 & 2/10\\
    \end{tabular}
    \caption{Success rates for our physical robot soccer experiments.}
\label{tab:results_physical}
\end{wraptable}

We further evaluate GuDA in a physical robot soccer task in which a NAO V6 robot must dribble a ball to the goal from the Easy and Hard initializations shown in Fig.~\ref{fig:init_1} and \ref{fig:init_2}.
The agent observes its position and orientation as well as the ball's position using vision-based state estimation.
The ball's dynamics depend on how the robot's feet contact the ball, and since foot positions are not observed, the ball's dynamics appear highly stochastic to the agent.
This stochasticity coupled with noisy state estimation makes this task notably difficult.
%
%
%
%
We collect data using a policy pre-trained in a low-fidelity soccer simulator with simplified dynamics and perfect state estimation (soccer-sim, Fig.~\ref{fig:guda_soccer_sim}).
Our dataset contains a single physical trajectory of the agent dribbling the ball from the center of the field to the goal (Fig.~\ref{fig:demo}).
This data is highly suboptimal for two reasons: (1) we trained the demonstrator in a low-fidelity simulator, and (2) the robot fumbled the ball and had to take an indirect route to the goal.
%
%


\begin{wrapfigure}{R}{0.5\linewidth}
    \centering
    \includegraphics[width=\linewidth]{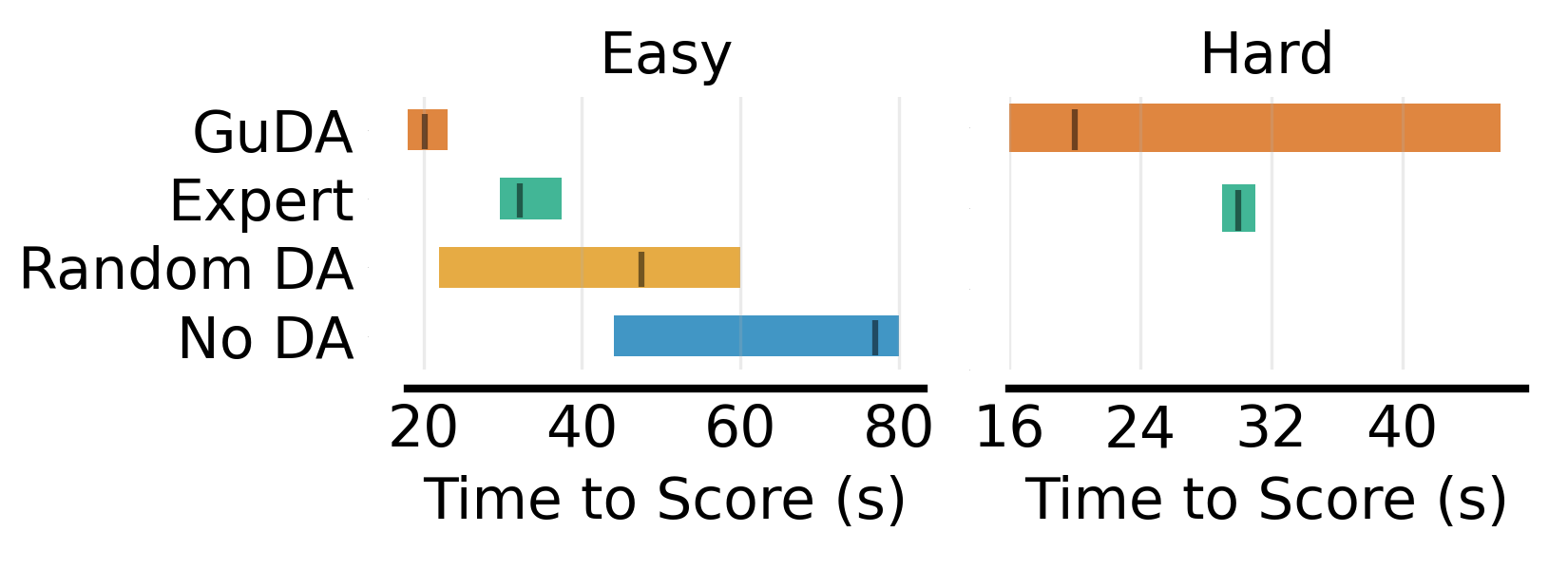}
    \caption{IQM time to score over 10 attempts with 95\% stratified bootstrap confidence intervals. Lower times are better.  GuDA's confidence interval in Hard is wide because of a single trial in which the agent scored after an unusually hard kick moved the ball to the opposite end of the field. Since MoCoDA failed to score in both tasks, we exclude it from this figure.}
    \label{fig:physical_time}
\end{wrapfigure}

To apply GuDA, we first identify two task-relevant behaviors in our initial dataset (Fig.~\ref{fig:demo}): the robot executing a tight turn to the ball ($\tau_1$), and the robot scoring with the ball away from the sideline ($\tau_2$).
We then define a sampling procedure to generate augmented trajectories that trace out the path an expert might take to successfully score (Fig.~\ref{fig:guda_demo}): we \texttt{Translate} and \texttt{Rotate} $\tau_1$ to demonstrate the agent approaching the ball at a favorable angle, and then we \texttt{Translate}, \texttt{Rotate}, and \texttt{Reflect} $\tau_2$ to demonstrate the agent scoring with the ball away from the sideline.
%

We generate 1 million augmented samples using GuDA, MoCoDA, and Random DA, and we train agents using IQL~\citep{Kostrikov2021OfflineRL} for 1 million policy updates. 
We also compare agents to the demonstrator we used to collect our physical trajectory.
%
%
Table~\ref{tab:results_physical} and Fig.~\ref{fig:physical_time} show the success rate and IQM time to score for each agent over 10 attempts at each initialization.
%
With the Easy initialization, GuDA scores faster and more frequently than MoCoDA, Random DA, and No DA. 
GuDA and the demonstrator policy have similar success rates, but GuDA scores significantly faster than the demonstrator as well.
We attribute this speedup to how the GuDA policy trained on augmented data that matches the physical world's dynamics (since our DAFs are dynamics-invariant) whereas our demonstrator policy trained in a low-fidelity simulator.
With the Hard initialization, only the GuDA agent can consistently score; MoCoDA, Random DA, No DA policies always kick the ball out of bounds.
Even the demonstrator policy almost always fails.
Our results show that GuDA not only outperforms MoCoDA and Random DA (\textbf{H2}) but also enables an agent to surpass its demonstrator in a difficult physical task with just a single suboptimal trajectory (\textbf{H1}).
%
%

%

\section{Conclusion}

In this work, we introduced Guided Data Augmentation (GuDA), a human-guided data augmentation (DA) framework that generates expert-quality augmented data without the expense of real-world task interaction.
%
In GuDA, a user imposes a series of simple rules on the DA process to automatically generate augmented samples that approximate expert behavior.
GuDA serves as a intuitive way to integrate human expertise into offline RL; instead of requiring that an expert demonstrate a near-optimal sequence of actions to solve a task, GuDA simply requires the user to understand what augmented data represents progress toward task completion.
Empirically, we demonstrate that GuDA outperforms a widely-applied random DA strategy as well as a model-based DA strategy and enables offline learning from a limited set of potentially suboptimal data.
Furthermore, we show how GuDA yields an effective policy in a physical robot soccer task when given a single highly suboptimal trajectory.
Our findings emphasize how a more intentional approach to DA can yield substantial performance gains.
%

The core limitation of GuDA is that it requires domain knowledge to specify sampling procedures. 
Since the sampling procedures required to generate expert augmented data are task dependent, GuDA must be implemented separately for each task.
In many navigation and object manipulation tasks, these rules can be derived from basic intuitions on what task progress looks like and are simple to implement.
However, GuDA is less applicable to tasks in which it is difficult to assess the quality of a trajectory segment (\textit{e.g.} chess). 
While our empirical analysis focuses on offline RL and behavior cloning, GuDA can in principle be applied to other learning methods -- both offline and online. 
Future work should study how GuDA interacts with other learning methods such as inverse RL and online RL.
Furthermore, a broader analysis investigating the the most effective way to integrate augmented data into offline RL -- similar to the analysis of~\cite{corrado2024understandingDA} for online RL -- would further strengthen the effectiveness of GuDA as well as other DA techniques.

\subsubsection*{Broader Impact Statement}
\label{sec:broaderImpact}

Our work focuses on fundamental RL research, and we thus see no direct negative societal consequences.
In this work, we propose a data augmentation framework (GuDA) that generates expert-quality augmented data and improves the performance of offline RL and behavior cloning methods. 
Since GuDA outperforms existing data augmentation methods on both simulated and physical tasks and yields effective policies even when given a small amount of suboptimal data, it can be applied to real-world tasks (where expert data is often scarce) and positively impact society.

%



\bibliography{main}
\bibliographystyle{rlc}

\appendix

\newpage
\section{Task Descriptions}
\label{app:task_descriptions}

\begin{figure}
    \centering
    \begin{subfigure}{0.32\linewidth}
        \includegraphics[width=\linewidth]{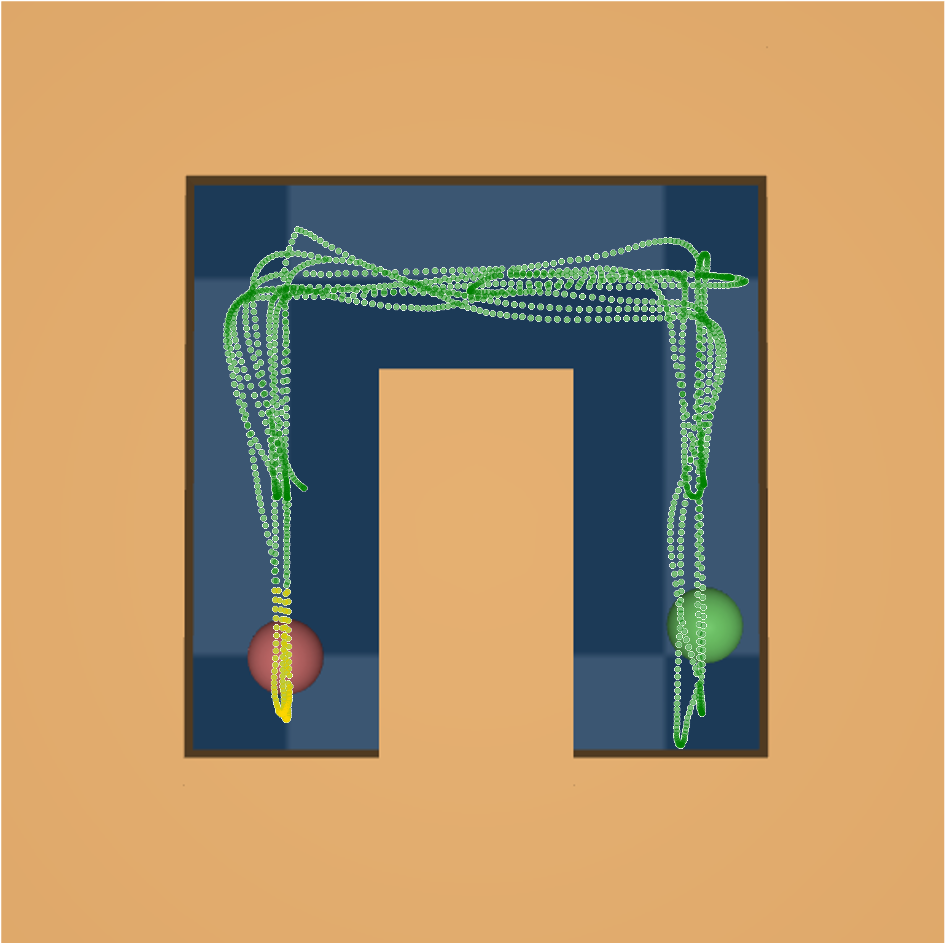}
        \caption{maze2d-umaze}
    \end{subfigure}
    \begin{subfigure}{0.32\linewidth}
        \includegraphics[width=\linewidth]{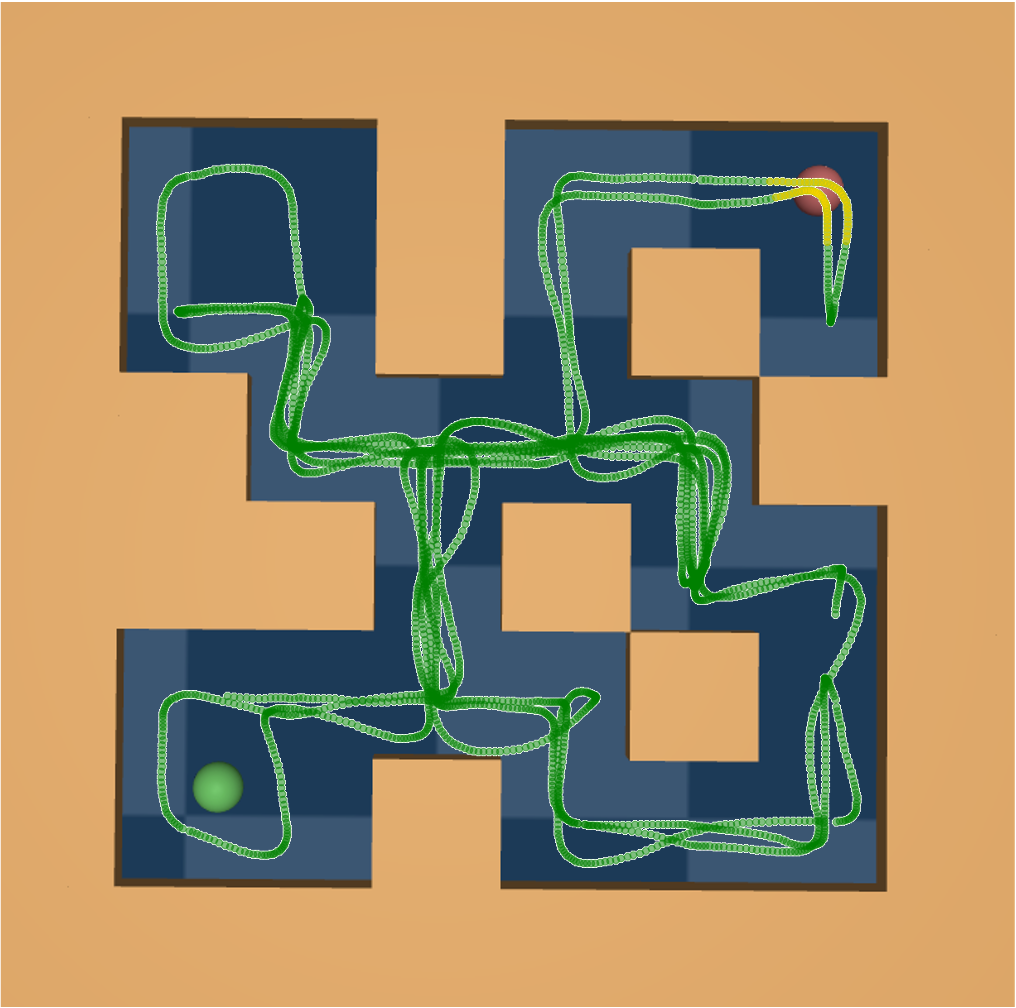}
        \caption{maze2d-medium}
    \end{subfigure}
        \begin{subfigure}{0.32\linewidth}
        \includegraphics[width=\linewidth]{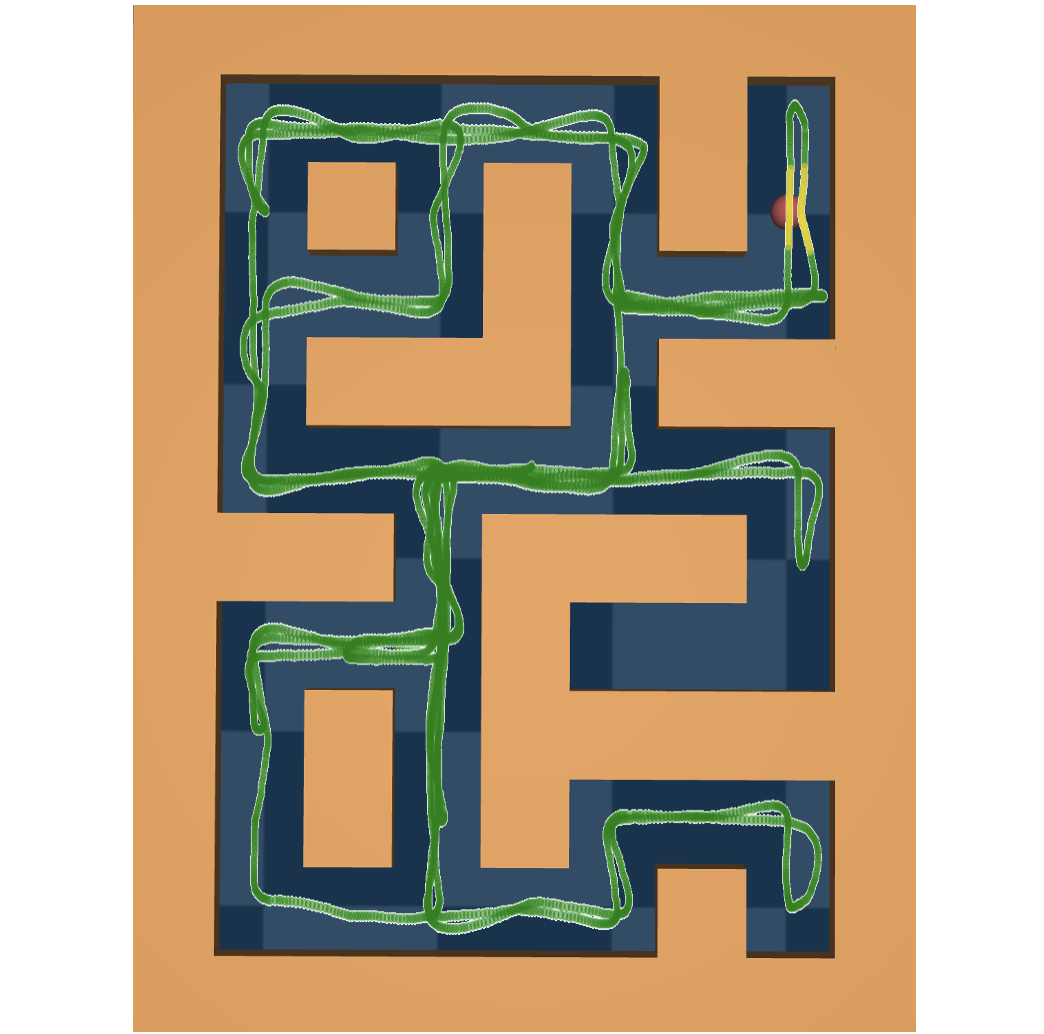}
        \caption{maze2d-large}
    \end{subfigure} 
    \\
    \begin{subfigure}{0.32\linewidth}
        \includegraphics[width=\linewidth]{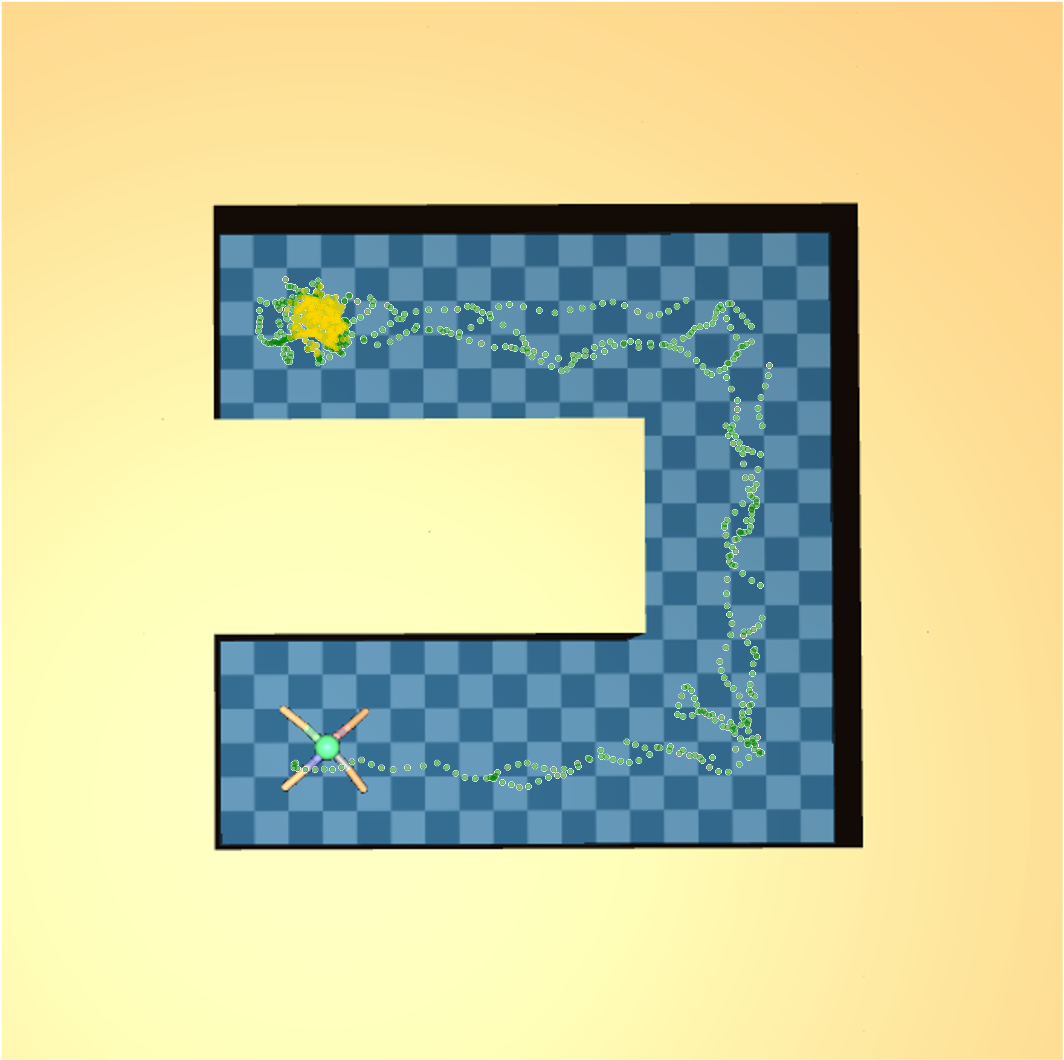}
        \caption{antmaze-umaze}
    \end{subfigure}
    \begin{subfigure}{0.32\linewidth}
        \includegraphics[width=\linewidth]{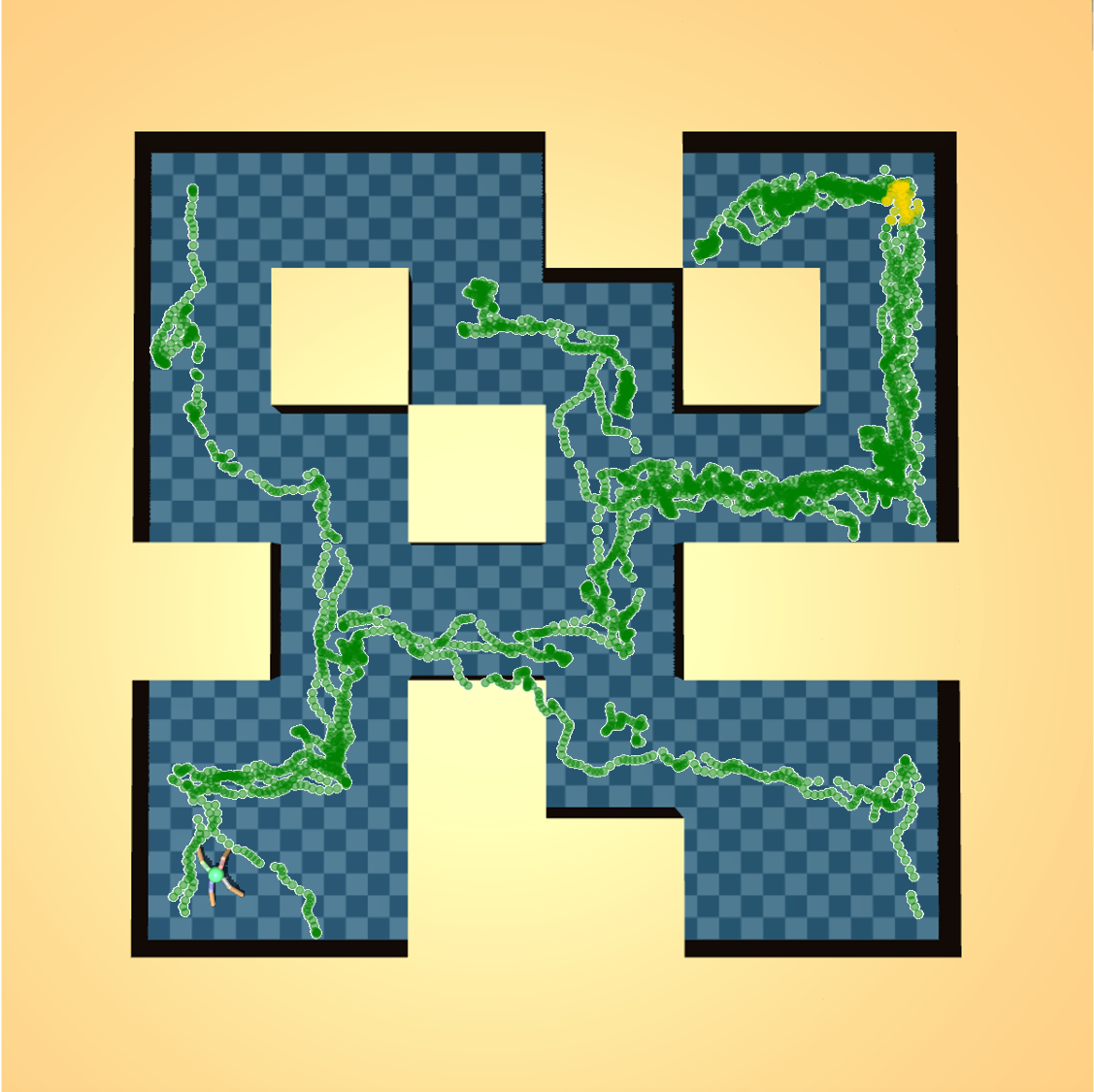}
        \caption{antmaze-medium}
    \end{subfigure}
        \begin{subfigure}{0.32\linewidth}
        \includegraphics[width=\linewidth]{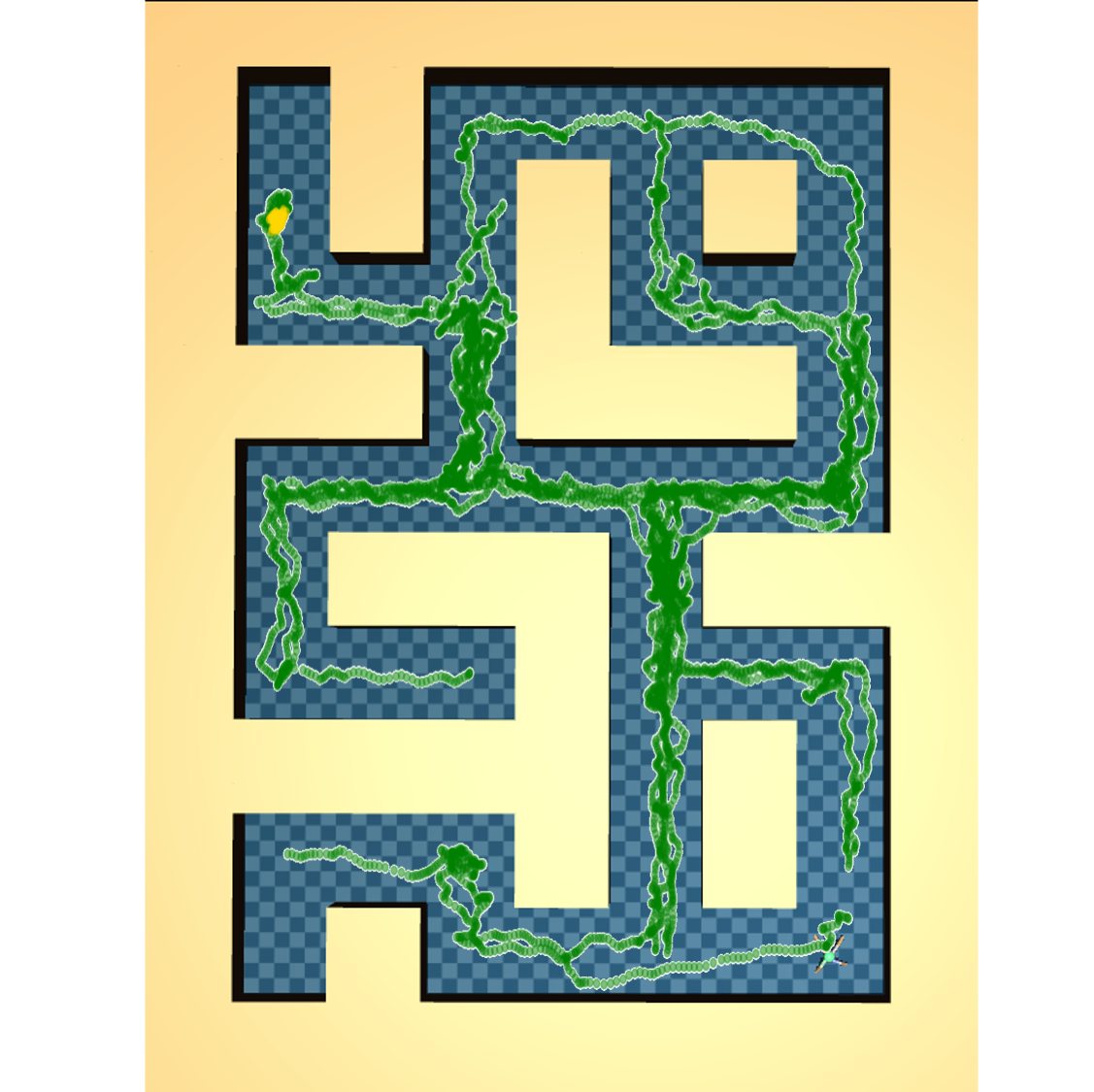}
        \caption{antmaze-large}
    \end{subfigure} 
    \\
    \begin{subfigure}{0.32\linewidth}
        \includegraphics[width=\linewidth]{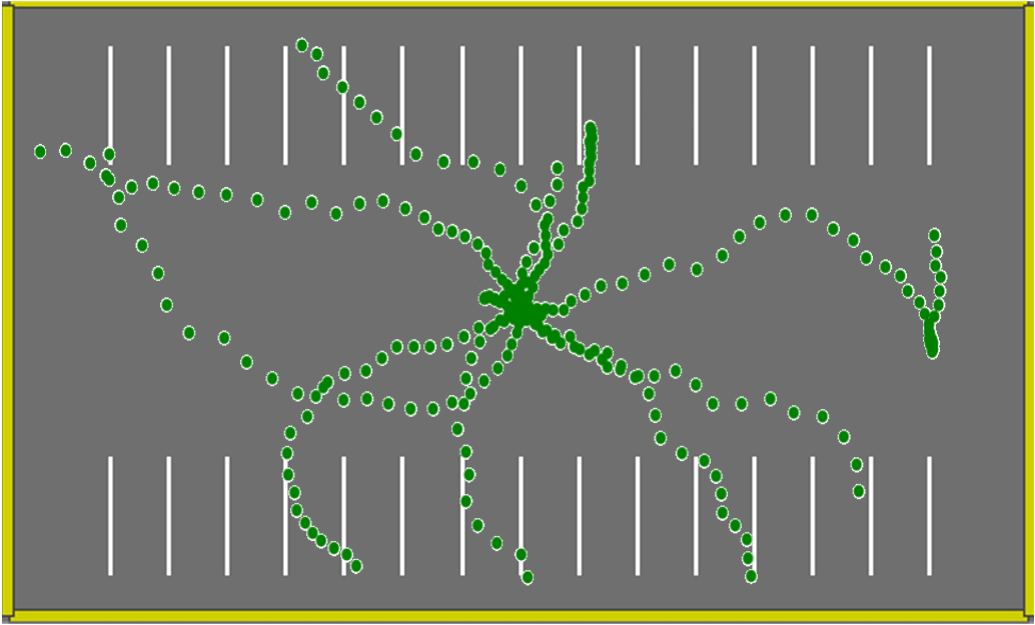}
        \caption{parking}
    \end{subfigure}
    \begin{subfigure}{0.32\linewidth}
        \includegraphics[width=\linewidth]{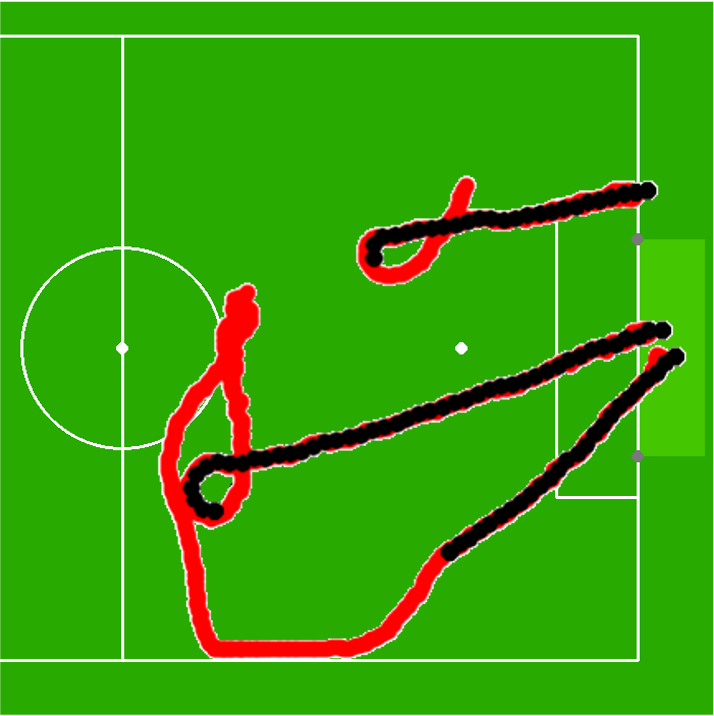}
        \caption{soccer-sim}
    \end{subfigure}
    \begin{subfigure}{0.32\linewidth}
        \includegraphics[width=\linewidth]{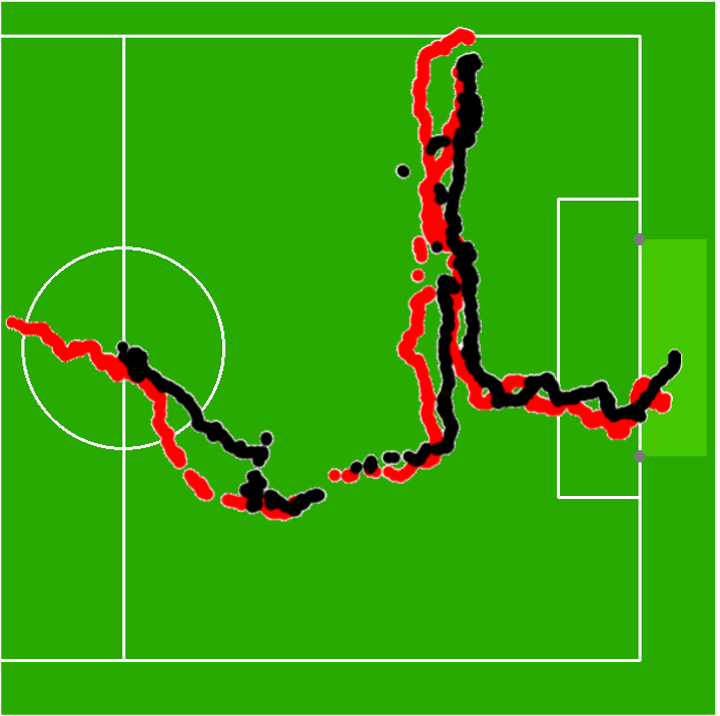}
        \caption{soccer-physical}
    \end{subfigure} 
    \caption{
    Visualizations of initial datasets. In maze2d and antmaze tasks, gold indicate data points for which the agent receives a nonzero reward. In soccer-sim and soccer-physical tasks, red denotes the agent and black denotes the ball.
    }
    \label{fig:initial_datasets}
\end{figure}

In this section, we describe each task in our empirical analysis. Fig.~\ref{fig:initial_datasets} visualizes the initial datasets used in each task.

\subsection{Maze2d} A force-actuated point-mass must navigate to a fixed goal from a random initial position.
The agent observes its position $(x, y)$ and velocity $(v_x, v_y)$, and the agent's actions take the form $\va = (f_x, f_y)$ where $f_x$ and $f_y$ are linear forces applied to the agent in the $x$ and $y$ directions, respectively. 
The agent receives +1 reward for being in positions within a disk of radius 0.5 centered at the the goal and 0 reward otherwise.

\subsection{Antmaze} This task is essentially the same as the maze2d task except the agent is replaced with a quadruped ``ant''. 
The agent must navigate to a fixed goal from a fixed initial maze position. 
The agent observes its position $(x, y)$, its height above the ground $z$, its orientation expressed as a quaternion, as well as the angle and angular velocities or all eight of its joints. 
The agent's action consists of torques to apply to each of the agent's joint.
The agent receives +1 reward for being in positions within a disk of radius 0.5 centered at the the goal and 0 reward otherwise. 

\subsection{Parking} 
An autonomous vehicle must park front-first into a designated parking spot. The agent observes it's current position $(x, y)$, velocity $(v_x, v_y)$, as well as the $\sin $ and $\cos$ of its heading $\theta$ (\textit{i.e.} the direction the front of the car is facing). 
The agent's state is thus 
$$\vs_\text{agent} = (x, y, v_x, v_y, \sin{\theta}, \cos{\theta}).$$
The agent also observes a goal $\vg$ consisting of the $(x_g, y_g)$ position of the parking spot, the desired velocity $(v_{g,x}, v_{g,y})$ at the parking spot (which is always set to $(0, 0)$), as well as the sine and cosine of car's desired heading $\theta_g$ at the parking spot (which is either $\theta_g = +\pi/2$ or $\theta_g = -\pi/2$) :
$$\vg = (x_g, y_g, 0, 0, \sin{\theta_g}, \cos{\theta_g}).$$    
The full state is $\vs = (\vs_\text{agent}, \vg)$.
%
The agent selects actions $\va = (a_\text{acc}, a_\text{steer})$ where $a_\text{acc}$ is the agent's acceleration in direction $\theta$ and $a_\text{steer}$ controls the agent's change in direction.

The agent receives a dense reward based on its distance to the parking spot and how closely the car aligns with the spot. 
If the agent crashes into one of the walls surrounding the parking lot, it receives a $-5$ reward penalty:

\begin{equation}
    r = -\sqrt{|\vs - \vg|} - 5\cdot \mathbbm{1}_\text{crash}
\end{equation}

\subsection{Soccer-sim} A robot (agent) must kick a ball to a fixed goal location. Robot and ball positions are initialized uniformly at random across the entire field. 
The observation contains the the following features:
\begin{itemize}
    \item  $(x_\text{robot to ball}, y_\text{robot to ball}) = (x_\text{robot} - x_\text{ball}, y_\text{robot} - y_\text{ball})$, the vector difference between the robot and ball positions.
    \item $(x_\text{ball to goal}, y_\text{ball to goal}) = (x_\text{ball} - x_\text{goal}, y_\text{ball} - y_\text{goal})$, the vector difference between the ball and goal positions.
    \item  $(\sin(\theta_\text{robot to ball}), \cos(\theta_\text{robot to ball}))$, where $\theta_\text{robot to ball}$ denotes the angle between the direction the robot is facing and the ball.
    \item  $(\sin(\theta_\text{ball to goal}), \cos(\theta_\text{ball to goal}))$, where $\theta_\text{ball to goal}$ denotes the angle between the ball and the goal.
\end{itemize}

The action $\va = (a_\theta, a_x, a_y)$ has three components: $a_\theta$ rotates the direction the robot is facing, and $(a_x, a_y)$ controls the robot's change in $(x, y)$ position.
The agent receives reward based on its distance to the ball and the ball's distance to the goal: 
    
\begin{equation}
    r = \begin{cases}
        \frac{0.9}{d_\text{agent to ball}} + \frac{0.1}{d_\text{ball to goal}} + \mathbbm{1}_\text{ball at goal}, & \text{if the agent is facing the ball} \\
        \mathbbm{1}_\text{ball at goal}, & \text{if the agent is not facing the ball}
    \end{cases}
\end{equation}

where $d_\text{agent to ball}$ is the Euclidean distance between the agent and the ball, $d_\text{ball to goal}$ is the Euclidean distance between the ball and the goal, and $\mathbbm{1}_\text{ball at goal}$ is an indicator function that returns $1$ when ball is at the goal and $0$ otherwise.
We say the agent is facing the ball if $|\theta_\text{robot to ball}| < 30^\circ $

\subsection{Soccer-physical} An agent must kick a ball to a fixed goal location. Agent and ball positions are initialized as shown in Fig.~\ref{fig:init_1} and Fig.~\ref{fig:init_2}. 
The agent receives reward based on its distance to the ball and the ball's distance to the goal. 
This task uses the same observation space, action space, and reward function used in the soccer-sim task.

\section{Sampling Procedures}
\label{app:sampling_rules}


\begin{figure*}
    \centering
    \begin{subfigure}{0.49\linewidth}
        \includegraphics[width=\linewidth]{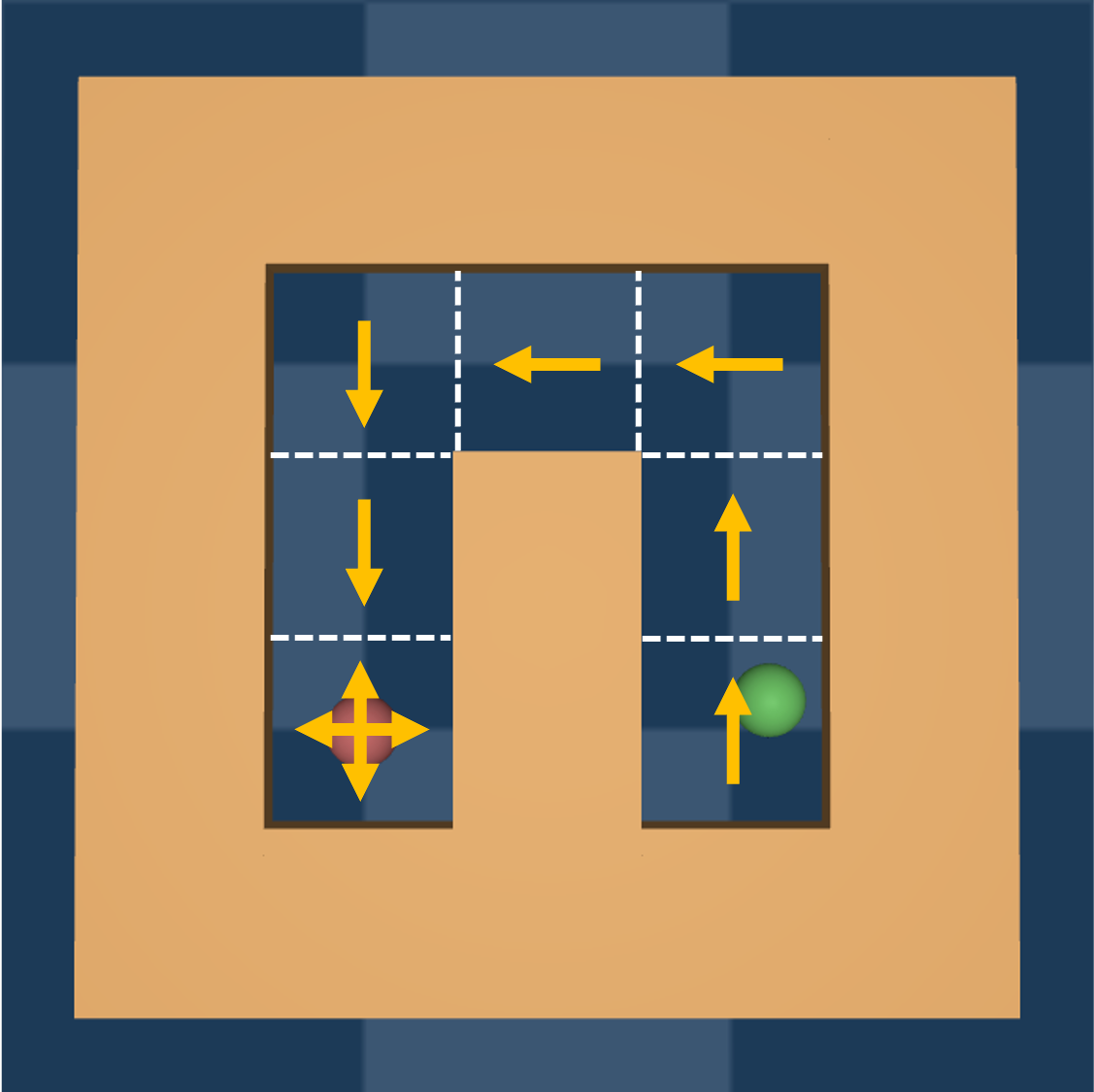}
        \caption{An illustration of the near-optimal displacement directions $\theta^*(\vx)$ in the maze2d-umaze-v1 task. Near the goal, (red ball), we sample rotation angles uniformly at random.}
        \label{fig:maze2d_angles}
    \end{subfigure}
    \hfill
    \begin{subfigure}{0.49\linewidth}
        \includegraphics[width=\linewidth]{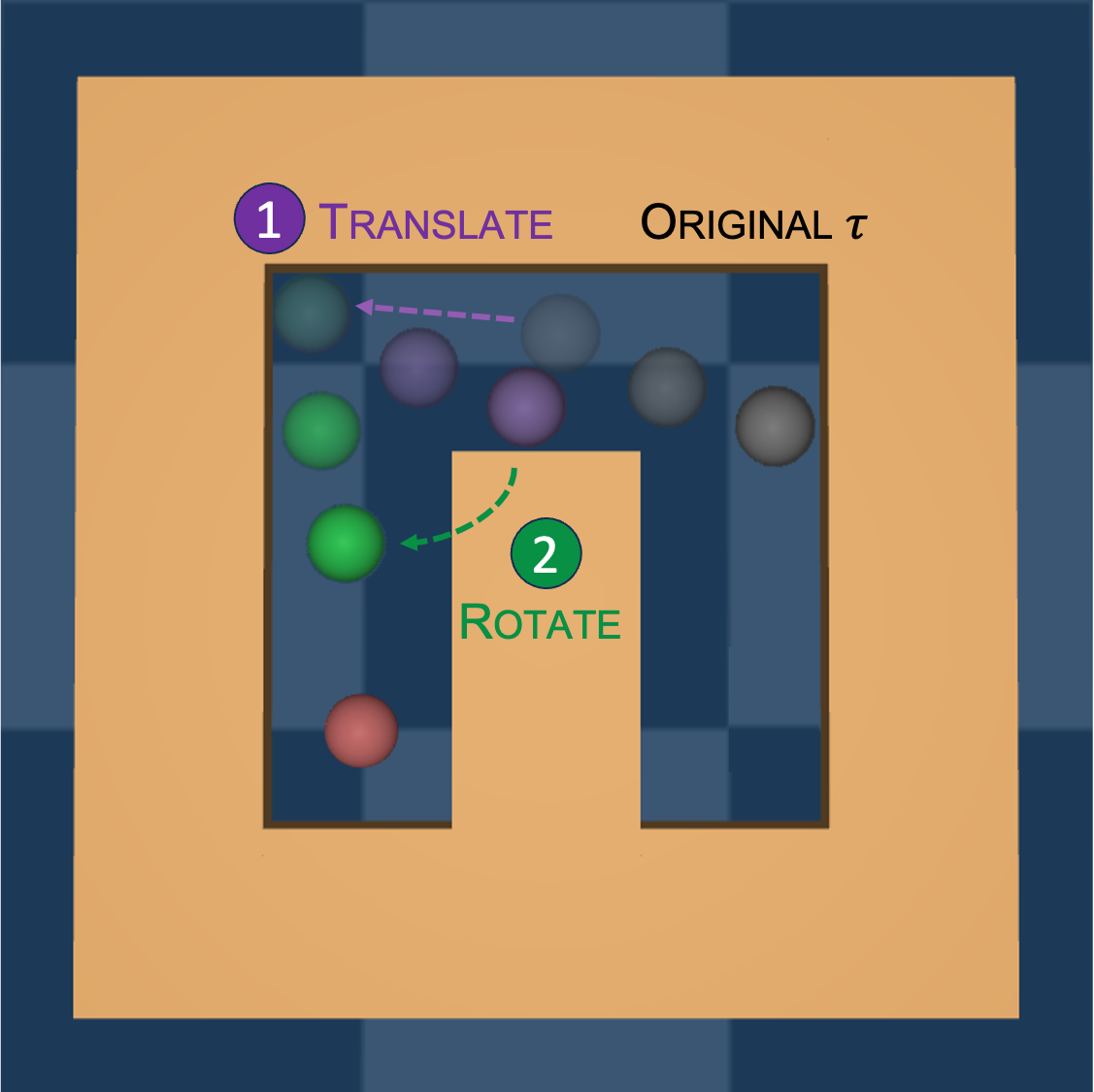}
        \caption{An illustration of the near-optimal displacement directions $\theta^*(\vx)$ in the maze2d-umaze-v1 task. Near the goal, (red ball), we sample rotation angles uniformly at random.}
        \label{fig:maze2d_procedure}
    \end{subfigure}
\end{figure*}

In this appendix, we provide a formal description of the sampling procedures we use to guide DA in our empirical analysis.
More concretely, we define distributions over translations $\sP(x,y|\tau)$, rotations $\Theta(\theta|\tau)$, and reflections $\sR(\tau)$ for each task.
We refer the reader to Table~\ref{tab:guided_augmentations} for a high-level description of guided data augmentations for each task.
In all descriptions, we let $\tau$ denote an input trajectory to be augmented.

\subsection{Maze2d}

In this task, we use the \texttt{Translate} and \texttt{Rotate} DAFs. 
Since the agent is initialized to a random position in the maze, an expert policy will visit all maze position. Thus, we let $\sP(x,y|\tau)$ be a uniform distribution over all valid maze positions for all $\tau$. 
We note that this distribution over maze positions is identical to the distribution used in a random DA strategy.

Before describing how we sample rotation angles, we introduce a few relevant quantities.
Let $(\Delta x, \Delta y)$ denote $\tau$'s displacement (the difference between $\tau$'s final and initial positions), and let $\theta(\tau) = \arctan{\frac{\Delta y}{\Delta x}}$ be the displacement angle of the original trajectory segment.
A user can easily compute $\theta(\tau)$ given $\tau$.
Additionally, let $\theta^*(x,y)$ be a function returning a near-optimal displacement direction at position $(x,y)$.
Fig.~\ref{fig:maze2d_angles} illustrates how we define $\theta^*(\tau)$ in the maze2d-umaze-v1 task. 
We divide the maze into cells (indicated by white dashed lines) and then label each cell with a desired displacement direction (arrows). 
%
%
Note that for the cell containing the goal, have no preferred displacement direction.

After translating the agent, we sample rotation angles for \texttt{Rotate} from  $\Theta(\theta|\tau) = (\theta^*(x,y) + \varepsilon) - \theta(\tau)$ where $(x,y)$ is the agents initial position in $\tau$, and $\varepsilon$ is a noise parameter distributed according to  $\textsf{Unif}([-\pi/6, +\pi/6])$.
We add noise to the rotation angle because offline RL methods learn to follow expert trajectories more effectively when expert data is noisy~\citep{KumarHSL22}.
Intuitively, $(\theta^*(x,y) + \varepsilon) - \theta(\tau)$ is the rotation angle required to rotate the agent's displacement direction from $\theta(\tau)$ to $(\theta^*(x,y) + \varepsilon)$.
A visualization of this sampling procedure can be found in Fig.~\ref{fig:maze2d_procedure}.

\begin{wrapfigure}{R}{0.4\linewidth}
\vspace{-1em}
    \centering
    \includegraphics[width=0.9\linewidth]{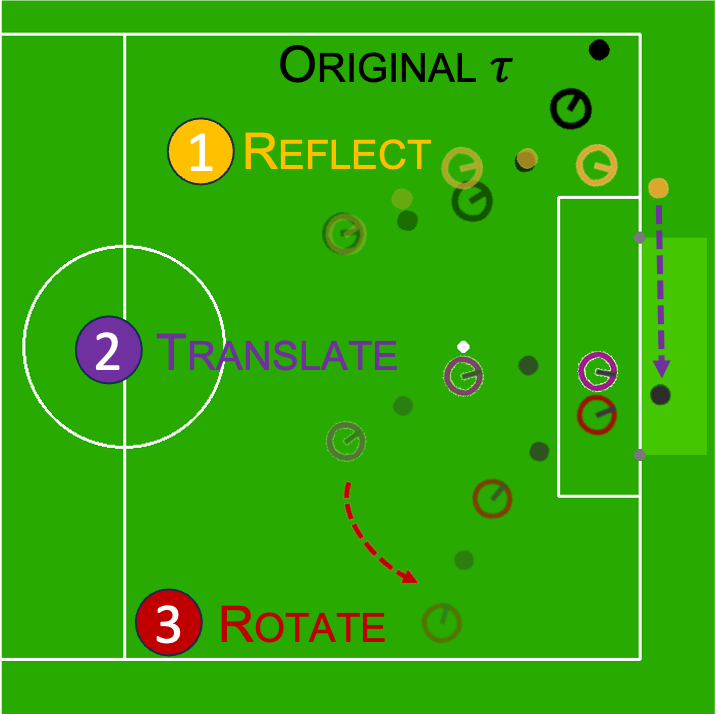}
    \caption{Visualization of the sampling procedure we use in soccer-sim.}
    \label{fig:soccer_procedure}
    \vspace{-3em}
\end{wrapfigure}

\subsection{Antmaze}

In this task, we use the \texttt{Translate} DAF.
Since the agent is initialized to a \textit{fixed} position in the maze, an expert policy will only visit a subset of maze position near the optimal path toward the goal. 
Thus, using the same notation established in the previous section for maze2d, we sample new positions from $\sP(x, y|\tau) = \textsf{Unif}(\{(x,y) : |\theta(\tau) - \theta^*(x,y)| < \frac{\pi}{4}, (x, y) \text{ is near the optimal path to the goal}\})$, a uniform distribution of task-relevant maze positions for which the agent's original displacement angle is within $\pi/4$ of the optimal displacement angle.
Similar to maze2d, we define $\theta^*(x,y)$ by dividing the maze into cells and then labeling each cell with a desired displacement angle. 
We then fetch all cells whose optimal displacement angle closely aligns with the agent's original displacement, and then randomly sample a new position from one of these cells.

\subsection{Parking}

In this task, we use the \texttt{Translate} and \texttt{Rotate} DAFs.
We first sample a new goal $\vg$ uniformly at random from $\sG$, and then translate $\tau$ such that its final position is a $\vg = (x_g, y_g)$, \textit{i.e.}, $\sP(x,y|\tau)$ places probability $1$ on the goal's position $(x_g, y_g)$. 

After translating, we rotate the agent such that the car's heading at the final step in $\tau$ is closely aligned with the parking spot's heading $\theta_g$.
We sample rotation angles from $\Theta(\theta|\tau) = (\theta_g + \varepsilon) - \theta(\tau)$, where $\varepsilon$ is a noise parameter distributed according to $\textsf{Unif}([-\pi/6, +\pi/6])$, and $\theta_\tau$ is heading angle of the last transitions in $\tau$.
Intuitively, $(\theta_g + \varepsilon) - \theta(\tau)$ is the rotation angle required to rotate agent's heading from $\theta(\tau)$ to $(\theta_g + \varepsilon)$.

\subsection{Soccer-sim}

In this task, we first reflect $\tau$ with probability $\sR(\tau) = 0.5$.
Then, we translate the $\tau$ so the ball's final position is at the goal.
Thus, we sample a new position from $\sP(x,y|\tau) = \textsf{Unif}(\{(x,y) : (x,y) \text{ is inside the goal}\})$.
Last, we rotate $\tau$ by a rotation angle sampled uniformly at random from while making sure the rotate trajectory remains in-bounds.
A visualization of this sampling procedure can be found in Fig.~\ref{fig:soccer_procedure}.



\section{MoCoDA Baseline}
\label{app:mocoda}

In this section, we provide additional details regarding MoCoDA experiments. 
We use the author's original implementation~\citealp[]{}{pitis2022mocoda}.

MoCoDA can in principle generate expert-quality augmented data if we specify a \textit{parent distribution} $P(\vs, \va)$ that is distributed according to the $(\vs, \va)$ distribution an expert might observe, but doing so requires us to explicitly describe the distribution of expert actions. 
We do not have access to this information.
However, it is nevertheless fairly simple to specify a distribution $P(\vs, \cdot)$ over task-relevant states.
In all tasks, we choose a parent distribution that is uniform over task-relevant agent positions, corresponding to MoCoDA-U in the original paper~\citep{pitis2022mocoda}.
At the implementation level, MoCoDA-U fits a Gaussian mixture model $P_\vtheta(\vs, \va)$ parameterized by $\vtheta$ to the provided dataset (while exploiting causal independence to generalize beyond the dataset's support). 
Then, this $P_\vtheta(\vs, \va)$ is reweighed to be uniform over agent positions and used as the parent distribution.
We note that this choice of parent distribution closely aligns with  our sampling procedures detailed in Appendix~\ref{app:sampling_rules}, which sample transformations uniformly at random over a small subset of task-relevant positions.

\section{Hyperparameter Tuning}
\label{app:hyperparams}
\begin{table}[]
    \centering
    \begin{tabular}{c|c|l}
        Algorithm & Hyperparameter & Values \\
        \hline
        \multirow{2}{*}{BC} & MLP network hidden layers & $(64, 64), (256, 256)$ \\
        & learning rate & $10^{-3}, 10^{-4}, 10^{-5}, 10^{-6}$  \\
        \hline
        \multirow{3}{*}{TD3+BC~\citep{fujimoto2021minimalist}} & MLP network hidden layers & $(64, 64), (256, 256)$ \\
         & actor/critic learning rates & $10^{-3}, 10^{-4}, 10^{-5}$  \\
        & $\alpha$ & $2.5, 5, 7.5, 10$ \\
        \hline
        \multirow{3}{*}{AWAC~\citep{nair2020awac}} & MLP network hidden layers & $(64, 64), (256, 256)$ \\
        & actor/critic learning rates & $10^{-3}, 10^{-4}, 10^{-5}$  \\
        & inverse advantage weight $\lambda$ & $0.5, 1, 2$ (and $0.1$ for antmaze) \\
        \hline
        \multirow{3}{*}{IQL~\citep{Kostrikov2021OfflineRL}} & MLP network hidden layers & $(64, 64)$ \\
        & actor/critic learning rates & $10^{-4}, 10^{-5}, 10^{-6}$  \\
        & inverse temperature $\beta$ & $1, 5, 10$ \\
        & expectile $\tau$ & $0.5, 0.7, 0.9$ \\
        \hline
    \end{tabular}
    \caption{Hyperparameter values we considered for each algorithm. 
    }
    \label{tab:hyperparams}
\end{table}

We tune all algorithms and DA strategies separately using a hyperparameter sweep over values listed in Table~\ref{tab:hyperparams}. 
In the main paper, we report the hyperparameters yielding the largest IQM return over 10 seeds.

Since it would be time-consuming and costly to evaluate all IQL hyperperameter settings in the physical robot soccer tasks, we first evaluate IQL hyperparameter settings in soccer-sim. 
We identify four IQL policies with the largest IQM return in soccer-sim, and then evaluate each of these four policies in the physical task.
We report results for the IQL policy yielding the largest success rate in both the Easy and Hard initializations.

\end{document}